\pdfoutput=1

\documentclass[11pt,table]{article}

\usepackage[preprint]{coling}

\usepackage{times}
\usepackage{latexsym}

\usepackage[T1]{fontenc}

\usepackage[utf8]{inputenc}

\usepackage{microtype}

\usepackage{inconsolata}

\usepackage{graphicx}

\usepackage{pgfplots, pgfplotstable}
\pgfplotsset{compat=1.15}

\usepackage{booktabs}
\usepackage{multirow}
\usepackage{fontawesome5}
\usepackage{adjustbox}
\usepackage{placeins}
\usepackage[inline]{enumitem}

\newcommand{\dialectname}[1]{\textbf{\MakeUppercase{#1}}}
\newcommand{\intent}{\multicolumn{1}{@{}r@{}}{\texttt{[intent]}}}

\newcommand{\munichdata}{\mbox{de-muc}} %
\newcommand{\deba}{\mbox{de-ba}}
\newcommand{\dest}{\mbox{de-st}}
\newcommand{\gsw}{gsw}
\newcommand{\multitask}{$\times$}
\newcommand{\sequential}{$\rightarrow$}

\newcommand{\xlmr}{\mbox{XLM-R}}

\title{Improving Dialectal Slot and Intent Detection with Auxiliary Tasks:\\A Multi-Dialectal Bavarian Case Study}

\author{Xaver Maria Krückl*\kern1pt\textsuperscript{\faMountain} 
\And 
Verena Blaschke*\kern1pt\textsuperscript{\faMountain\kern1pt\faRobot} \\
\textsuperscript{\faMountain} MaiNLP, Center for Information and Language Processing, LMU Munich, Germany \\
\textsuperscript{\faRobot} Munich Center for Machine Learning (MCML), Munich, Germany \\
{\tt xaver.krueckl@gmail.com, \{verena.blaschke, b.plank\}@lmu.de} 
\And 
Barbara Plank\kern1pt\textsuperscript{\faMountain\kern1pt\faRobot}}

\definecolor{gold}{RGB}{212, 163, 40}
\definecolor{darkgreen}{RGB}{37, 125, 65}
\definecolor{darkred}{RGB}{161, 13, 23}

\definecolor{colEn}{RGB}{24, 25, 66}
\definecolor{colDe}{RGB}{28, 50, 148}
\definecolor{colDeBa}{RGB}{151, 56, 168}
\definecolor{colDeSt}{RGB}{217, 113, 205}
\definecolor{colDeBy}{RGB}{212, 87, 93}
\definecolor{colGsw}{RGB}{252, 186, 3}

\definecolor{eng}{RGB}{224, 224, 224}
\definecolor{bar}{RGB}{235, 138, 106}

\newcommand{\gold}[1]{\textcolor{gold}{#1}}
\newcommand{\predcorrect}[1]{\textcolor{darkgreen}{#1~\faCheck}}
\newcommand{\predwrong}[1]{\textcolor{darkred}{#1~\raisebox{-1pt}{\faTimes}}}
\newcommand{\gloss}[1]{\small\textcolor{gray}{\textit{#1}}}

\newcommand{\colourcircle}[1]{\raisebox{-1pt}{\textcolor{#1}{\faCircle}}}

\newlength{\shortrow}
\newcommand\blfootnote[1]{%
  \begingroup
  \renewcommand\thefootnote{}\footnote{#1}%
  \addtocounter{footnote}{-1}%
  \endgroup
}

\begin{document}
\maketitle
\begin{abstract}
Reliable slot and intent detection (SID) is crucial in natural language understanding for applications like digital assistants. 
Encoder-only transformer models fine-tuned on high-resource languages generally perform well on SID. 
However, they struggle with dialectal data, where no standardized form exists and training data is scarce and costly to produce. 
We explore zero-shot transfer learning for SID, focusing on multiple Bavarian dialects, for which we release a new dataset for the Munich dialect.
We evaluate models trained on auxiliary tasks in Bavarian, and compare joint multi-task learning with intermediate-task training.
We also compare three types of auxiliary tasks: token-level syntactic tasks, named entity recognition (NER), and language modelling.
We find that the included auxiliary tasks have a more positive effect on slot filling than intent classification (with NER having the most positive effect), and that intermediate-task training yields more consistent performance gains.
Our best-performing approach improves intent classification performance on Bavarian dialects by 5.1 and slot filling F1 by 8.4~percentage points.
\end{abstract}

\section{Introduction}
\label{sec:introduction}

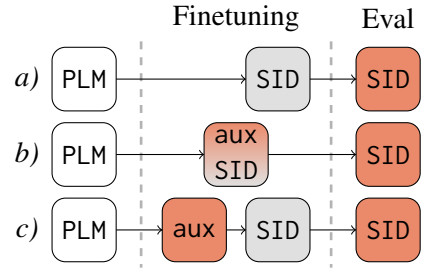
\begin{figure}
    \centering
    \begin{tikzpicture}[
    align=left,
    mynode/.style={draw, inner sep=2pt, outer sep=0pt, minimum size=24pt, rounded corners=5pt, font=\ttfamily},
    bavarian/.style={fill=bar},
    english/.style={fill=eng},
    mydashed/.style={dashed, black!30, line width=1pt},
    myenum/.style={font=\itshape},
]
\node[mynode] at (0, 2) (PLM1) {PLM};
\node[mynode] at (0, 0) (PLM2) {PLM};
\node[mynode] at (0, 1) (PLM3) {PLM};
\node[mynode, english] at (2.55, 2) (SID1) {SID};
\node[mynode, bavarian] at (1.45, 0) (AUX) {aux};
\node[mynode, english] at (2.55, 0) (SID2) {SID};
\node[mynode, top color=bar, bottom color=eng] at (2, 1) (AUXSID) {aux\\SID};
\node[mynode, bavarian] at (4, 2) (EVAL1) {SID};
\node[mynode, bavarian] at (4, 0) (EVAL2) {SID};
\node[mynode, bavarian] at (4, 1) (EVAL3) {SID};
\draw[mydashed] (0.75, 2.5) -- (0.75, -0.5);
\draw[mydashed] (3.25, 2.5) -- (3.25, -0.5);
\draw[->] (PLM1) -- (SID1);
\draw[->] (SID1) -- (EVAL1);
\draw[->] (PLM2) -- (AUX);
\draw[->] (AUX) -- (SID2);
\draw[->] (SID2) -- (EVAL2);
\draw[->] (PLM3) -- (AUXSID);
\draw[->] (AUXSID) -- (EVAL3);
\node[] at (2, 2.8) {Finetuning};
\node[] at (4, 2.8) {Eval};
\node[myenum] at (-0.75, 2) {a)};
\node[myenum] at (-0.75, 1) {b)};
\node[myenum] at (-0.75, 0) {c)};
\end{tikzpicture}
    \caption{\textbf{Overview of evaluated setups.}
    We fine-tune pre-trained language models (PLMs) on English SID data (grey~\colourcircle{eng}) and evaluate them on Bavarian (red~\colourcircle{bar}).
    We compare multiple setups:
    \textit{a)}~no auxiliary tasks,
    \textit{b)}~multi-task learning by jointly training on English SID data and Bavarian auxiliary tasks (``aux''),
    \textit{c)}~intermediate-task training on Bavarian, then fine-tuning on English SID data.
    }
    \label{fig:methodolopgy}
\end{figure}

\blfootnote{*Equal contribution.}%
Most research on natural language processing (NLP) for digital assistants has focused on standardized languages, despite the large degree of dialectal variation exhibited by many languages and the positive attitude towards dialectal versions of such technologies expressed by some speaker communities \citep{blaschke-etal-2024-dialect}. 

A core task of natural language understanding (NLU) is to detect the intent of an input to a digital assistant (e.g., the instruction ``delete all alarms'' belongs to the \textit{cancel alarm} class) and to tag it for specific slots (e.g., ``all'' should be tagged as the \textit{reference} associated with the intent).
However, classifying dialectal inputs is still challenging as contemporary models are less proficient due to the scarcity of low-resource and especially dialectal training data \citep{Zampieri_Nakov_Scherrer_2020}. 
To overcome this issue, transferring task knowledge cross-lingually from high-resource language data to low-resource varieties is a strategy widely used in NLU (\citealp{upandhyay_atis_zeroshot, schuster-etal-2019-cross-lingual, xu-etal-2020-end}, inter alia).
While many approaches have focused on cross-lingual transfer via embedding transmission and machine translation, \citet{van-der-goot-etal-2021-masked} use non-English auxiliary task data for zero-shot transfer to other languages.

Inspired by this setup and by intermediate-task training procedures \cite{pruksachatkun-etal-2020-intermediate}, we use auxiliary tasks to analyze and improve zero-shot transfer learning for slot and intent detection (SID) for Bavarian dialects (Figure~\ref{fig:methodolopgy}).
To account for intra-dialectal variation, we evaluate on two previously released Bavarian datasets and introduce a third test set.
For the auxiliary tasks, we use three recent Bavarian datasets for syntactic annotations, named entity recognition (NER), and masked language modelling (MLM).

\noindent
We make the following contributions:
\begin{itemize}
\item
We release a new Bavarian slot and intent detection evaluation dataset~(\S\ref{sec:data-munich}).\footnote{To be included in \url{https://github.com/mainlp/xsid}.}
\item
We examine how training on auxiliary NLP tasks in Bavarian affects SID performance~(\S\ref{sec:results-multitask-sequential}). We compare both the integration of the auxiliary tasks into the training setup (joint multi-task learning vs.\ intermediate-task training) and the tasks themselves.
\item
To analyze the robustness of the results, we examine performance and data differences between the dialectal test sets~(\S\ref{sec:results-language-comparison}, \ref{sec:results-bavarian-differences}) and include additional datasets~(\S\ref{sec:results-other-sid}).
\end{itemize}

\noindent
We share our code publicly.\footnote{\url{https://github.com/mainlp/auxtasks-bavarian-sid}}

\section{Related Work}
\label{sec:related-work}

\paragraph{Slot and intent detection for dialects and non-standard varieties}
\label{sec:related-work-sid-dialects}
Research on SID for low-resource languages, including non-standard and dialectal varieties, has started receiving more attention.
This trend starts with \citet{van-der-goot-etal-2021-masked}, who introduce a multilingual SID dataset, xSID, containing South Tyrolean, a Bavarian dialect (more details in \S\ref{sec:data-xsid}).
xSID has since been extended with dialectal data from Upper Bavaria \citep{winkler-etal-2024-slot}, data in Bernese Swiss German and Neapolitan \citep{aepli-etal-2023-findings}, and eight Norwegian dialects \citep{maehlum-scherrer-2024-nomusic}.

Similarly to our study, \citet{van-der-goot-etal-2021-masked} experiment with multi-task learning, although they only have Standard German auxiliary data at their disposal for the South Tyrolean test data.
Other approaches focus on tokenization issues or data augmentation.
\citet{srivastava-chiang-2023-fine} tackle tokenization issues caused by spelling differences by injecting character-level noise into standard-language training data, which improves the performance on the dialectal test sets.
\citet{munoz-ortiz-etal-2025-evaluating} find that encoding text with visual representations (rather than ones based on subword tokens) improves transfer from Standard German to German dialects for intent classification.
\citet{abboud-oz-2024-towards} fine-tune a masked language model on dialectal data to generate synthetic training data for German and Arabic dialects.
\citet{malaysha-etal-2024-arafinnlp} organized a shared task on intent detection in four Arabic dialects, where the top systems all involve model ensembling and translating the training data into the test dialects \cite{ramadan-etal-2024-ma, elkordi-etal-2024-alexunlp24, fares-touileb-2024-babelbot}.

In the context of spoken intent classification, other work focuses on variation in spoken Italian \cite{koudounas23_interspeech} and English \cite{gerz-etal-2021-multilingual, rajaa2022skits2iindianaccentedspeech, he-garner-2023-interpreter}.

\paragraph{Multi-task learning (MTL)}
Joint MTL learning involves jointly training a model on several tasks.
\citet{ruder2017overviewmultitasklearningdeep} provides a general overview.
\citet{martinez-alonso-plank-2017-multitask} find that tasks with non-skewed label distributions lend themselves best as auxiliary tasks for sequence tagging.
\citet{schroder-biemann-2020-estimating} show that auxiliary tasks which are more similar to the target tasks result in better target performance.

Regarding MTL for SID,
\citet{wang2021encoding} train a transformer model on dependency parsing, POS tagging, and SID, with different layers attending to the different tasks. 
They find that the syntactic tasks improve SID performance (especially when both are included), and that jointly producing slot and intent labels is also beneficial.

\Citet{van-der-goot-etal-2021-masked} use English training data for SID but additionally exploit non-English auxiliary task data, hypothesizing that this helps their models to learn additional linguistic properties of the target language.
They find syntactic tasks to be useful for slot filling for one pre-trained language model but not another, and harmful for intent detection. Similarly, they find masked language modelling (MLM) to be of use for slot filling but not intent classification. Machine translation as auxiliary task yielded worse performance.

\paragraph{Intermediate-task training}
While MTL is about fine-tuning a model \textit{simultaneously} on multiple tasks, intermediate-task training concerns first fine-tuning a model on one or more auxiliary tasks and \textit{subsequently} fine-tuning it on the target task.
In a similar vein to some MTL results, 
\citet{poth-etal-2021-pre} and \citet{padmakumar-etal-2022-exploring} find the similarity between the intermediate and target task to be important.
Similarly, \citet{pruksachatkun-etal-2020-intermediate} evaluate models on inference and reading understanding tasks and find including intermediate tasks also related to reasoning to be useful.
\citet{padmakumar-etal-2022-exploring} further find that including multiple intermediate tasks at once often yields better results than only including one, although the interactions of tasks are difficult to predict.

In the context of cross-lingual evaluation, \citet{louvan2021investigating} find that continued pre-training via target-language MLM has mixed results. %
\citet{phang-etal-2020-english} show that even in cross-lingual scenarios, intermediate-task learning on the source language can be beneficial.

Some recent studies include both MTL and intermediate-task training.
\citet{weller-etal-2022-use} find that MTL with several auxiliary tasks tends to perform worse than with just one additional task, and that MTL beats intermediate-task training when the target task has less data than the auxiliary task.
\citet{montariol-etal-2022-multilingual} focus on cross-lingual hate speech detection and add auxiliary tasks in multiple languages (including the target language). They find joint MTL setups to outperform intermediary task training, and semantic auxiliary tasks to be more beneficial than syntactic ones.

\section{Background: Bavarian Dialects}
\label{sec:background-bavarian}

\begin{figure}
    \centering
    \includegraphics[width=\linewidth]{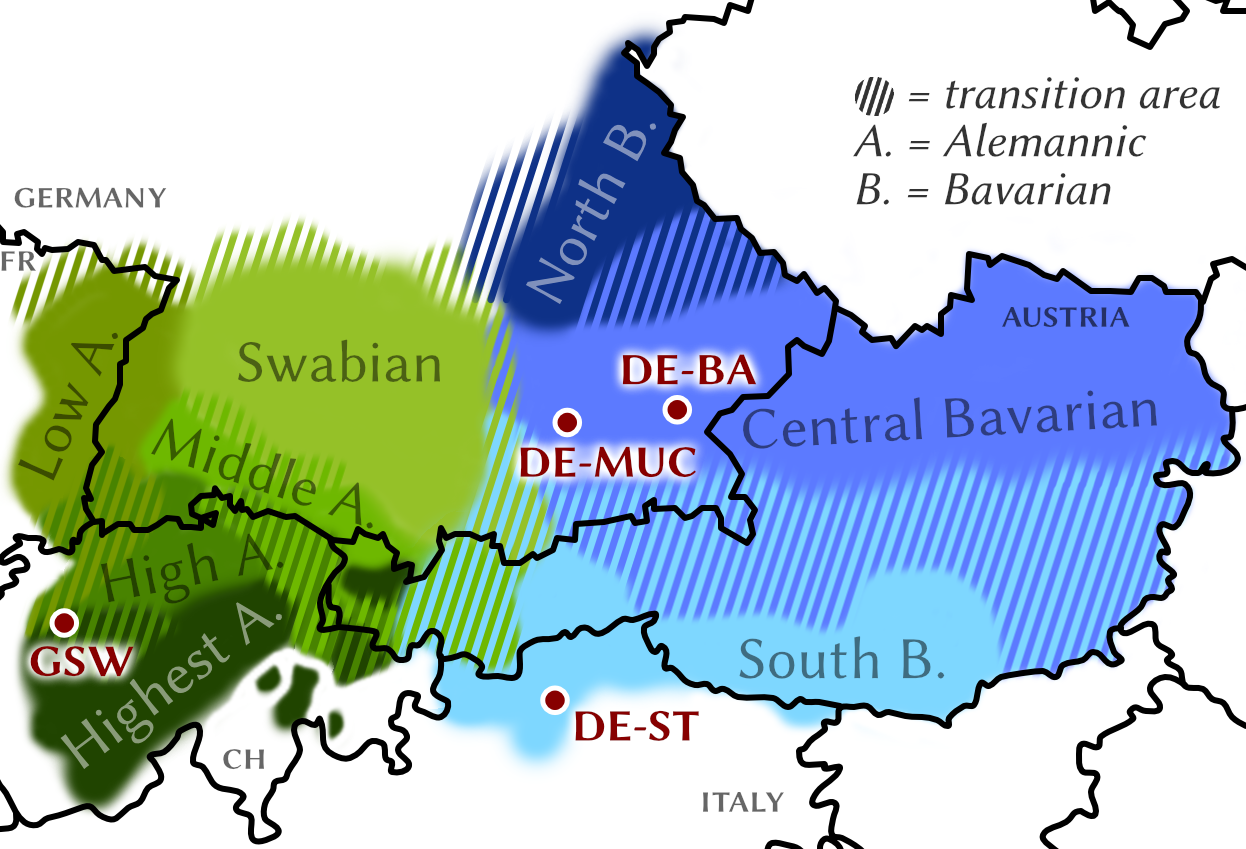}
    \caption{\textbf{The Upper German dialect groups Bavarian} (blue, right) \textbf{and Alemannic} (green, left), based on \citet{wiesinger1983deutschedialekte}. The red dots show the xSID datasets included in this study and our new dataset, \munichdata. 
    }
    \label{fig:map}
\end{figure}

Bavarian dialects differ from Standard German in phonetics, phonology, word choice, and morphosyntax \cite{merkle1993bairische}.
There is no established orthography or standard variety of Bavarian.
The Bavarian dialects belong to the Upper German dialect group and are split into three major subgroups (Northern, Central, and Southern Bavarian; Figure~\ref{fig:map}), mostly based on sound differences \cite{wiesinger1983deutschedialekte}.
There is also phonetic/\allowbreak{}phonological and lexical variation \textit{within} these groups \cite[passim]{rowley2023boarisch}.
The pronunciation differences are also reflected in the spelling choices made in the different training and evaluation datasets in our study, although the spellings also reflect idiosyncratic preferences. %
We compare the Bavarian SID test sets in~\S\ref{sec:results-bavarian-differences}.

Some of the morphosyntactic differences between Bavarian and Standard German (cf.\ \citealp{blaschke-etal-2024-maibaam}) are relevant for SID, and recent work \cite{artemova-etal-2024-exploring} has shown that slot filling performance in German is negatively affected by dialectal syntactic structures.
Person names are typically preceded by definite articles, and the given name generally follows the family name \cite[pp.~69--71]{weiss1998syntax} -- this has been analyzed in the context of NER \cite{peng-etal-2024-sebastian} and might also be relevant for slot filling.
Furthermore, many NLU queries contain infinitive constructions of the form ``remind me to [do X]''. Such cases are often expressed with a nominalized infinitive construction (\citealp{bayer1993zum, bayer2004klitisiertes-zu}; see, e.g., Table~\ref{tab:reminder-get-paper}) that does not exist in Standard German.

Additionally, as in many other German dialects \cite{weise1910stundenbezeichnungen}, temporal expressions (relevant for \texttt{datetime} slots) can be expressed in ways that are not grammatical in Standard German, e.g., \textit{fia fünfe heid auf Nacht} ``for 5\textsc{pm} tonight'' (lit.\ ``for five today at night'') or \textit{um 3 nammiddog} ``at 3\textsc{pm}'' (lit.\ ``at 3 afternoon'').

\section{Data}
\label{sec:data}

\subsection{Slot and Intent Detection Data}
\label{sec:data-xsid}

\paragraph{xSID}
We use xSID~0.5 (\citealp{van-der-goot-etal-2021-masked}; \href{https://github.com/mainlp/xsid/blob/main/LICENSE}{CC~BY-SA~4.0}), which provides development and test sets (300 and 500 sentences, respectively) for slot and intent detection in a range of languages, as well as a large English training set (44k sentences).
It covers 16 intents and 33 different slot types.
The data consist of re-annotated English sentences from SNIPS \citep{coucke2018snips} and a Facebook dataset \citep{schuster-etal-2019-cross}. The non-English development and test splits are translations.

xSID~0.5 contains multiple Upper German dialects (Figure~\ref{fig:map}), none of which are standardized: South Bavarian as spoken in South Tyrol (\textbf{\dest{}}; included in the first xSID release), Central Bavarian as spoken in Upper Bavaria (\textbf{\deba{}}; \citealp{winkler-etal-2024-slot}), and Swiss German as spoken in Bern (\textbf{gsw}; \citealp{aepli-etal-2023-findings}).
We focus on the Bavarian test sets, but include the Swiss German data as well as the Standard German (\textbf{de}) and English (\textbf{en}) test sets in an additional evaluation~(\S\ref{sec:results-language-comparison}).

\paragraph{Munich Bavarian evaluation data}
\label{sec:data-munich}

To investigate the effect of intra-dialectal variation and different translation choices, we create a second Central Bavarian translation.
The new test and development set is in the dialect spoken in Munich (\textbf{\munichdata}), translated by a native speaker (one of the authors).
The translation is directly from English, without referencing either the Standard German or dialectal versions, as was also done for the other dialect translations.
The (sentence-level) intent labels are the same as in English and the other languages; the (token-level) slot spans were annotated by the translator.
As there is no Bavarian orthography, \munichdata{} represents the spelling preferences of the translator. 
The grapheme--phoneme mapping is similar to that of Standard German and reflects the translator's pronunciation.
Most words are lower-cased, also nouns that would be capitalized in German.
Named entities are left untranslated and, per the xSID guidelines, grammatical mistakes in the original sentences are also adopted in the translations.

Our Munich Bavarian translations are the most similar to the other Central Bavarian ones (\deba) on a word and character level (see Appendix~\ref{sec:appendix-dataset-distances}).

We share a data statement \cite{bender-friedman-2018-data} in Appendix~\ref{sec:appendix-data-statement}.

\paragraph{Additional evaluation data}
To evaluate whether some of our findings generalize to other Bavarian datasets, we use test sets provided by \citet{winkler-etal-2024-slot}.
They collected naturalistic data by asking Bavarian speakers to come up with queries for a digital assistant that match xSID's intents, and translated a subset of MASSIVE \cite{fitzgerald-etal-2023-massive} with the labels mapped to match xSID's.
The translator for MASSIVE is the same as for xSID's \deba{} set, and the contributors to the naturalistic data also come from the same region.

\subsection{Auxiliary Task Data Sets}
\label{sec:data-aux}

We use three Bavarian datasets for auxiliary NLP tasks.
These tasks are similar to ones explored in related work on MTL for SID~(\S\ref{sec:related-work}) and are additionally motivated by data availability.

\paragraph{Syntactic dependencies and POS tags (UD)}
As token-level information and linguistic structure might be useful for slot annotations, we include two syntactic tasks: dependency parsing and part-of-speech (POS) tagging.
The Universal Dependencies~v2.14 (UD; \citealp{de-marneffe-etal-2021-universal}) treebank MaiBaam (\citealp{blaschke-etal-2024-maibaam}; \href{https://github.com/UniversalDependencies/UD_Bavarian-MaiBaam/blob/master/LICENSE.txt}{CC~BY-SA~4.0}) provides such dependency annotations and POS tags for Bavarian dialects from all three Bavarian dialect groups, including varieties spoken in South Tyrol, Upper Bavaria, and Munich.
MaiBaam contains some sentences from xSID, which we exclude from our experiments, leaving 975 sentences that we randomly divide into training and development data using a 90:10 split.

\paragraph{Named entity recognition (NER)}
Similarly to slot filling, NER concerns identifying and labelling spans of tokens as a sequence tagging task. 
BarNER~1.0 \citep{peng-etal-2024-sebastian} provides such annotations for named entities in Wikipedia articles (\href{https://creativecommons.org/licenses/by-sa/4.0/deed.en}{CC~BY-SA~4.0}) and tweets. 
Based on the inspection of a small data sample, \citeauthor{peng-etal-2024-sebastian} state that the most represented Bavarian dialect group is Central Bavarian (to which both \deba{} and \munichdata{} belong).
We use the predefined training and development splits (9k and 918 sentences, respectively), and use the fine-grained label set.

\paragraph{Masked language modelling (MLM)}
We also include MLM, as it is a common pre-training objective.\footnote{We note however that mDeBERTa~v.3 is pre-trained on replaced token detection rather than MLM \cite{he2021debertav3}.}
We use a subset of the Bavarian Wikipedia (1.5k~sentences, divided into 90\% training and 10\% development data), as pre-processed by \citet{artemova-plank-2023-low}.

\section{Methodology}
\label{sec:methodology}

We fine-tune pre-trained language models (PLMs) on xSID's English training data using MaChAmp 0.4.2 (commit \texttt{9f5a6ce}; \citealp{van-der-goot-etal-2021-massive}) with the same hyperparameters as \citet{van-der-goot-etal-2021-masked} did for their SID experiments. 

We evaluate slot predictions with strict slot F1, intent predictions with accuracy, and also calculate the proportion of sentences with fully correct predictions.
We treat SID itself as a multi-task setup as we jointly predict the slots and intent labels, and treat slot detection as a basic sequence labelling task with a final softmax layer.
We use the following task types for MaChAmp \cite{van-der-goot-etal-2021-massive}: 
\texttt{seq} (slot filling, NER, POS tagging),
\texttt{classification} (intent classification),
\texttt{mlm} (MLM), and
\texttt{dependency} (dependency parsing).
The loss for each task is weighted equally.
We use MaChAmp's default loss functions (cross-entropy loss for all tasks except dependency parsing, which uses negative log likelihood).
We provide mean scores across three runs for each experiment.

\noindent
We compare three types of experimental setups (Figure~\ref{fig:methodolopgy}):

\paragraph{Baseline}
We compare four commonly used PLMs, which we finetune on SID data without auxiliary tasks: the monolingual German GBERT \cite{chan-etal-2020-germans}, and the multilingual models mBERT \cite{devlin-etal-2019-bert}, \xlmr{} \cite{conneau-etal-2020-unsupervised}, and mDeBERTa~v.3 \cite{he2021debertav3, he2021deberta}.

Notably, mBERT's pretraining data also includes the Bavarian Wikipedia, which contains articles in all three of our test dialects.
\xlmr{} and mDeBERTa were pre-trained on the CC-100 dataset \cite{conneau-etal-2020-unsupervised}, which does not contain Bavarian data. GBERT's pretraining data is in Standard German.
To limit computation costs, we use the base-sized versions.\footnote{We use 
\href{https://huggingface.co/deepset/gbert-base}{deepset/\allowbreak{}gbert-\allowbreak{}base} (license: \href{https://opensource.org/license/mit}{MIT}),
\href{https://huggingface.co/google-bert/bert-base-multilingual-cased}{google-bert/\allowbreak{}bert-\allowbreak{}base-\allowbreak{}multilingual-\allowbreak{}cased} (\href{https://choosealicense.com/licenses/apache-2.0/}{Apache~2.0}),
\href{https://huggingface.co/FacebookAI/xlm-roberta-base}{FacebookAI/\allowbreak{}xlm-\allowbreak{}roberta-\allowbreak{}base} (\href{https://opensource.org/license/mit}{MIT}),
and
\href{https://huggingface.co/microsoft/mdeberta-v3-base}{microsoft/\allowbreak{}mdeberta-\allowbreak{}v3-\allowbreak{}base} (\href{https://opensource.org/license/mit}{MIT}).
}
In the remaining setups, we only use mDeBERTa because of its strong performance as a baseline PLM~(\S\ref{sec:results-baselines}).

\paragraph{Multi-Task Learning}
We train the model to jointly predict labels for SID and at least one auxiliary task. %
We use \multitask{} to denote these setups, e.g., NER\multitask{}SID refers to training a model to simultaneously predict named entity labels, slots and intents.

\paragraph{Intermediate-Task Training}
We first train the model to predict labels for an auxiliary task, remove the task-specific head, optionally repeat this for a second auxiliary task, and then finally train the model to predict SID labels.
We use \sequential{} to denote these setups, e.g. NER\sequential{}SID refers to first training a model on NER data, then on SID data.
As a special case, we train some models first jointly on auxiliary tasks and then afterwards on SID (e.g., MLM\multitask{}NER\sequential{}SID).

\vspace{\baselineskip}\noindent
We apply each auxiliary task dataset to both fine-tuning setups.
For the settings with multiple auxiliary tasks, we select combinations that appear promising based on the results already obtained.
We were not able to examine all possible combinations due to computational restraints.

\section{Results and Analysis}
\label{sec:results}

\begin{figure}
    \centering
    \adjustbox{max width=\linewidth}{\newcommand{\intentcol}{teal!85}
\newcommand{\slotcol}{cyan!85}
\newcommand{\fullcol}{olive!75}

\trimbox{5mm 1.2mm 17.5mm 2.7mm}{
\begin{tikzpicture}[remember picture]
\pgfplotstableread{
Label                                             Slots    Intents  {Fully correct} StdevSlots StdevIntents StdevFull
{SID\\(mBERT\rlap{)}}                             42.9     64.3     11.3     1.6     2.9     1.6
{SID\\(XLM-R\rlap{)}}                             34.0     65.1      8.2     1.7     7.0     2.4
{SID\\(GBERT\rlap{)}}                             47.2     70.3     15.4     1.9     3.1     1.6
{SID\\(mDeBERTa\rlap{)}}                          45.3     73.5     15.1     2.0     6.6     2.5
UD\sequential{}SID                                49.3     73.8     19.0     2.2     7.3     2.2
UD\multitask{}SID                                 42.1     48.9     13.0     4.5     7.6     2.8
NER\sequential{}SID                               53.1     76.6     22.0     2.7     6.8     3.1
NER\multitask{}SID                                53.8     76.2     21.2     2.0     7.3     3.6
MLM\sequential{}SID                               39.6     71.8     12.1     3.8     7.1     3.4
MLM\multitask{}SID                                44.6     71.9     14.3     2.5     7.6     2.7
UD\sequential{}NER\\\sequential{}SID              54.3     78.3     22.6     3.3     6.4     3.2
UD\multitask{}NER\\\sequential{}SID               53.7     78.4     22.3     1.9     5.8     2.9
UD\multitask{}NER\\\multitask{}SID                44.7     62.6     15.1     4.6     6.2     2.5
MLM\multitask{}UD\\\sequential{}SID               49.2     74.0     19.3     2.7     6.0     2.1
MLM\sequential{}NER\\\sequential{}SID             51.5     76.8     21.1     2.2     7.4     2.0
MLM\multitask{}NER\\\sequential{}SID              53.7     78.6     22.9     2.3     7.2     2.7
MLM\multitask{}NER\\\multitask{}SID               54.8     77.9     22.3     1.7     7.5     2.7
MLM\multitask{}UD\\\multitask{}NER\multitask{}SID 48.7     58.0     17.0     2.4     7.9     2.4
}\datatable

\pgfplotsset{
    /pgfplots/xbar legend/.style={
        legend image code/.code={
            \draw [#1, draw=none] (0pt,-3pt) rectangle (12pt,4pt);
        },
    }
}
\begin{axis}[
    xbar,
    xmin = 0,
    height = 21cm,
    width = 8cm,
    ytick = data,
    xtick = \empty,
    yticklabels from table = {\datatable}{Label},
    y dir = reverse,
    tickwidth = 0pt,
    y axis line style = {opacity = 0},
    nodes near coords,
    nodes near coords style = {white, font=\scriptsize,
        shift={(axis direction cs:-3.2mm-\offset,0)},
        /pgf/number format/.cd, fixed zerofill, precision=1},
    enlarge y limits  = 0.04,
    enlarge x limits  = 0.02,
    xmajorgrids = false,
    tick label style={font=\footnotesize, inner sep=0pt, outer sep=0pt, align=right},
    legend columns = -1,
    legend style={
            at={(xticklabel cs:.5)},
            anchor=north,
            /tikz/every even column/.append style={
                column sep=.5cm,
            },
        },
    reverse legend,
    legend style={draw=none, font=\footnotesize,
                  /tikz/every even column/.append style={column sep=1.5mm}},
    barstyle/.style = {
        draw=none,
        bar width=7pt,
    },
]
\draw [\intentcol, dashed, line width=1pt] (73.5,-0.4) -- (73.5,17.0);
\draw [\slotcol, dashed, line width=1pt] (45.3,-0.1) -- (45.3,17.2);
\draw [\fullcol, dashed, line width=1pt] (15.1,0.2) -- (15.1,17.5);

\addplot [visualization depends on={\thisrow{StdevFull} \as \offset}, fill=\fullcol, barstyle, error bars/.cd, x explicit, x dir=both,] table [x={Fully correct}, y expr=\coordindex,
                x error plus expr=\thisrow{StdevFull},
                x error minus expr=\thisrow{StdevFull},] {\datatable};
\addplot [visualization depends on={\thisrow{StdevSlots} \as \offset}, fill=\slotcol, barstyle, error bars/.cd, x explicit, x dir=both,]
                table [x=Slots, y expr=\coordindex,
                x error plus expr=\thisrow{StdevSlots},
                x error minus expr=\thisrow{StdevSlots},] {\datatable};
\addplot [visualization depends on={\thisrow{StdevIntents} \as \offset}, fill=\intentcol, barstyle, error bars/.cd, x explicit, x dir=both,]
                table [x=Intents, y expr=\coordindex,
                x error plus expr=\thisrow{StdevIntents},
                x error minus expr=\thisrow{StdevIntents},] {\datatable};

\coordinate (z) at (74,-0.5);

\legend{Fully correct\hspace{15mm}\phantom{.}, Slots (strict F1), Intents (acc.)}
\end{axis}
\end{tikzpicture}
\begin{tikzpicture}[remember picture, overlay]
    \fill[white, transparency group, opacity=.5] (z) rectangle (-10,16.52);
\end{tikzpicture}
}}
    \caption{\textbf{Slot and intent detection results for the different models, in \%.}
    The results are averaged over the three Bavarian dialect test sets and three random seeds (standard deviations shown as error bars).
    Mean scores and standard deviations per individual dialect are in Appendix~\ref{sec:appendix-details}.
    The dashed lines denote the scores of the baseline model (no auxiliary tasks).
    The setups with auxiliary tasks also use mDeBERTa.
    The three pale entries at the top are worse-performing baseline models with alternative PLMs.
    }
\label{fig:overview_all_results_dialects}
\end{figure}
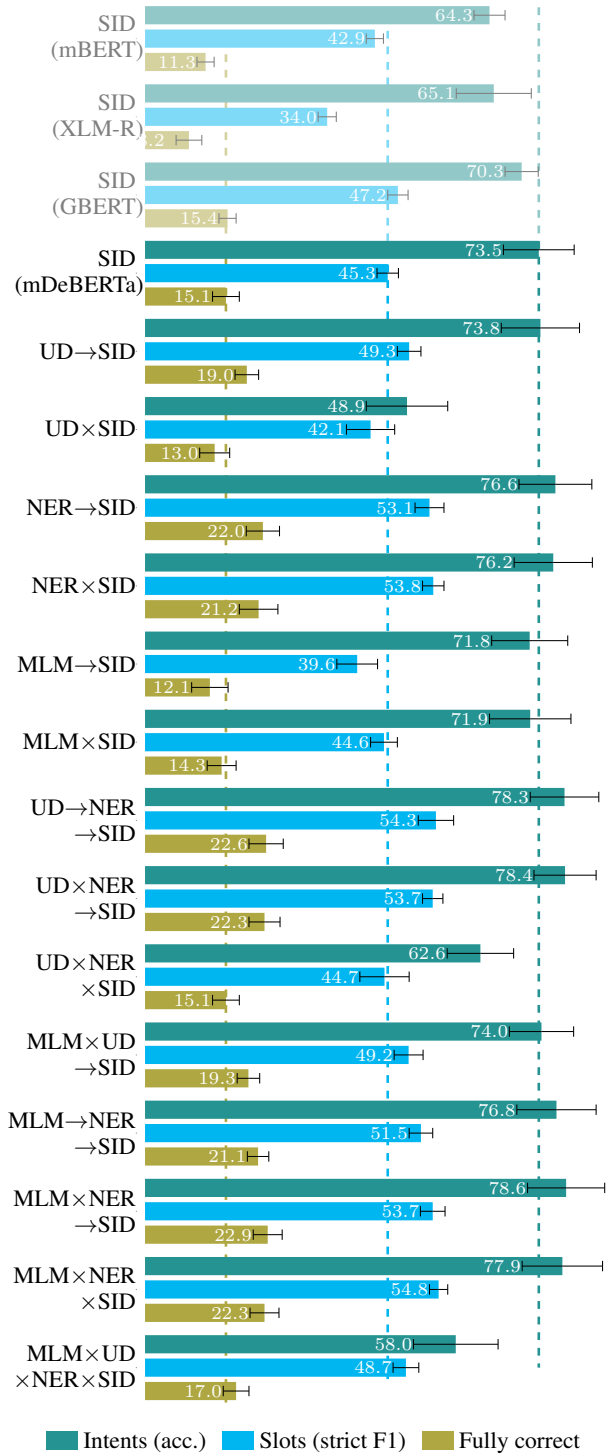

\newcommand{\spacer}{}
\newcommand{\spacerB}{\phantom{0}}
\begin{table*}[t]
\centering
\adjustbox{max width=\textwidth}{%
\setlength{\tabcolsep}{2pt}
\begin{tabular}{@{}l@{\hspace{8pt}}rrrrrrrrr@{\hspace{15pt}}rrrrrrrr@{}}
\toprule
& \multicolumn{8}{c}{Intents} & & \multicolumn{8}{c}{Slots}\\ \cmidrule(lr){2-9} \cmidrule(l){11-18}
& &  & \multicolumn{6}{c}{$\Delta$ to baseline} &  &  &  & \multicolumn{6}{c}{$\Delta$ to baseline} \\ \cmidrule(lr){3-9} \cmidrule(l){12-18} 
& \multirow{-2}{*}{Avg} &  & \multicolumn{1}{c}{ITT} & \multicolumn{1}{c}{MTL} &  & \multicolumn{1}{c}{UD} & \multicolumn{1}{c}{NER} & \multicolumn{1}{c}{MLM} &  & \multirow{-2}{*}{Avg} &  & \multicolumn{1}{c}{ITT} & \multicolumn{1}{c}{MTL} &  & \multicolumn{1}{c}{UD} & \multicolumn{1}{c}{NER} & \multicolumn{1}{c}{MLM} \\[-4pt]
&&& \small (\sequential{}SID) & \small (\multitask{}SID) &&&&&&&& \small (\sequential{}SID) & \small (\multitask{}SID)\\
\midrule
SID (mDeBERTa) & 73.5 &  & &  &  &  & & &  & 45.3 &  & &  &  &  & & \\
\midrule
UD\sequential{}SID & 73.8 &  & \cellcolor[HTML]{FDFFFE}+\spacer{}0.3 &  &  & {\cellcolor[HTML]{FDFFFE}+\spacer{}0.3} & & &  & 49.3 &  & \cellcolor[HTML]{E5F5ED}+\spacer{}3.9 &  &  & {\cellcolor[HTML]{E5F5ED}+\spacer{}3.9} & & \\
UD\multitask{}SID & 48.9 &  & & {\cellcolor[HTML]{E67E75}--24.6} &  & {\cellcolor[HTML]{E67E75}--24.6} & & &  & 42.1 &  & & {\cellcolor[HTML]{FBEEEC}--\spacer{}3.2} &  & {\cellcolor[HTML]{FBEEEC}--\spacer{}3.2} & & \\
NER\sequential{}SID & 76.5 &  & \cellcolor[HTML]{EBF7F1}+\spacer{}3.0 &  &  &  & \cellcolor[HTML]{EBF7F1}+\spacer{}3.0 & &  & 53.1 &  & \cellcolor[HTML]{CBEADB}+\spacer{}7.8 &  &  &  & \cellcolor[HTML]{CBEADB}+\spacer{}7.8 & \\
NER\multitask{}SID & 76.2 &  & & {\cellcolor[HTML]{EDF8F3}+\spacer{}2.7} &  &  & \cellcolor[HTML]{EDF8F3}+\spacer{}2.7 & &  & 53.8 &  & & {\cellcolor[HTML]{C7E9D8}+\spacer{}8.4} &  &  & \cellcolor[HTML]{C7E9D8}+\spacer{}8.4 & \\
MLM\sequential{}SID & 71.8 &  & \cellcolor[HTML]{FDF5F5}--\spacer{}1.8 &  &  &  & & \cellcolor[HTML]{FDF5F5}--\spacer{}1.8 &  & 39.6 &  & \cellcolor[HTML]{F9E0DE}--\spacer{}5.8 &  &  &  & & \cellcolor[HTML]{F9E0DE}--\spacer{}5.8 \\
MLM\multitask{}SID & 71.9 &  & & {\cellcolor[HTML]{FDF6F5}--\spacer{}1.6} &  &  & & \cellcolor[HTML]{FDF6F5}--\spacer{}1.6 &  & 44.6 &  & & {\cellcolor[HTML]{FEFBFA}--\spacer{}0.7} &  &  & & \cellcolor[HTML]{FEFBFA}--\spacer{}0.7 \\
UD\sequential{}NER\sequential{}SID & 78.3 &  & \cellcolor[HTML]{DFF3E9}+\spacer{}4.8 &  &  & {\cellcolor[HTML]{DFF3E9}+\spacer{}4.8} & \cellcolor[HTML]{DFF3E9}+\spacer{}4.8 & &  & 54.3 &  & \cellcolor[HTML]{C3E7D5}+\spacer{}9.0 &  &  & {\cellcolor[HTML]{C3E7D5}+\spacer{}9.0} & \cellcolor[HTML]{C3E7D5}+\spacer{}9.0 & \\
UD\multitask{}NER\sequential{}SID & 78.4 &  & \cellcolor[HTML]{DFF2E9}+\spacer{}4.8 &  &  & {\cellcolor[HTML]{DFF2E9}+\spacer{}4.8} & \cellcolor[HTML]{DFF2E9}+\spacer{}4.8 & &  & 53.7 &  & \cellcolor[HTML]{C7E9D8}+\spacer{}8.4 &  &  & {\cellcolor[HTML]{C7E9D8}+\spacer{}8.4} & \cellcolor[HTML]{C7E9D8}+\spacer{}8.4 & \\
UD\multitask{}NER\multitask{}SID & 62.6 &  & & {\cellcolor[HTML]{F4C5C1}--10.9} &  & {\cellcolor[HTML]{F4C5C1}--10.9} & \cellcolor[HTML]{F4C5C1}--10.9 & &  & 44.7 &  & & {\cellcolor[HTML]{FEFBFB}--\spacer{}0.6} &  & {\cellcolor[HTML]{FEFBFB}--\spacer{}0.6} & \cellcolor[HTML]{FEFBFB}--\spacer{}0.6 & \\
MLM\multitask{}UD\sequential{}SID & 73.9 &  & \cellcolor[HTML]{FDFEFE}+\spacer{}0.4 &  &  & \cellcolor[HTML]{FDFEFE}+\spacer{}0.4 & & \cellcolor[HTML]{FDFEFE}+\spacer{}0.4 &  & 49.2 &  & \cellcolor[HTML]{E6F5EE}+\spacer{}3.8 &  &  & \cellcolor[HTML]{E6F5EE}+\spacer{}3.8 & & \cellcolor[HTML]{E6F5EE}+\spacer{}3.8 \\
MLM\sequential{}NER\sequential{}SID & 76.8 &  & \cellcolor[HTML]{E9F7F0}+\spacer{}3.3 &  &  &  & \cellcolor[HTML]{E9F7F0}+\spacer{}3.3 & \cellcolor[HTML]{E9F7F0}+\spacer{}3.3 &  & 51.5 &  & \cellcolor[HTML]{D6EFE3}+\spacer{}6.2 &  &  &  & \cellcolor[HTML]{D6EFE3}+\spacer{}6.2 & \cellcolor[HTML]{D6EFE3}+\spacer{}6.2 \\
MLM\multitask{}NER\sequential{}SID & 78.6 &  & \cellcolor[HTML]{DDF2E8}+\spacer{}5.1 &  &  &  & \cellcolor[HTML]{DDF2E8}+\spacer{}5.1 & \cellcolor[HTML]{DDF2E8}+\spacer{}5.1 &  & 53.7 &  & \cellcolor[HTML]{C7E9D8}+\spacer{}8.4 &  &  &  & \cellcolor[HTML]{C7E9D8}+\spacer{}8.4 & \cellcolor[HTML]{C7E9D8}+\spacer{}8.4 \\
MLM\multitask{}NER\multitask{}SID & 77.9 &  & & {\cellcolor[HTML]{E2F4EB}+\spacer{}4.3} &  &  & \cellcolor[HTML]{E2F4EB}+\spacer{}4.3 & \cellcolor[HTML]{E2F4EB}+\spacer{}4.3 &  & 54.8 &  & & {\cellcolor[HTML]{C0E6D3}+\spacer{}9.5} &  &  & \cellcolor[HTML]{C0E6D3}+\spacer{}9.5 & \cellcolor[HTML]{C0E6D3}+\spacer{}9.5 \\
MLM\multitask{}UD\multitask{}NER\multitask{}SID & 58.0 &  &  & {\cellcolor[HTML]{EFADA7}--15.6} &  & {\cellcolor[HTML]{EFADA7}--15.6} & \cellcolor[HTML]{EFADA7}--15.6 & \cellcolor[HTML]{EFADA7}--15.6 &  & 48.7 &  & & {\cellcolor[HTML]{E9F6F0}+\spacer{}3.3} &  & \cellcolor[HTML]{E9F6F0}+\spacer{}3.3 & \cellcolor[HTML]{E9F6F0}+\spacer{}3.3 & \cellcolor[HTML]{E9F6F0}+\spacer{}3.3 \\
\midrule
Mean & &  & \cellcolor[HTML]{EFF9F4}+\spacer{}2.5 & {\cellcolor[HTML]{F7D7D4}--\spacer{}7.6} &  & {\cellcolor[HTML]{F9E0DE}--5.8} & \cellcolor[HTML]{FEFFFF}+\spacer{}0.2 & \cellcolor[HTML]{FEFAFA}--\spacer{}0.8 &  & &  & \cellcolor[HTML]{DCF1E7}+\spacer{}5.2 & {\cellcolor[HTML]{EDF8F2}+\spacer{}2.8} &  & {\cellcolor[HTML]{E8F6EF}+\spacer{}3.5} & \cellcolor[HTML]{D2EDE0}+\spacer{}6.7 & \cellcolor[HTML]{E8F6EF}+\spacer{}3.5 \\
Std. deviation & &  & \spacerB\spacerB{}2.6 & \spacerB{}{11.4} &  & \spacerB{}{11.4} & \spacerB\spacerB{}7.7 & \spacerB\spacerB{}7.1 &  & &  & \spacerB\spacerB{}4.9 & \spacerB\spacerB{}{5.2} &  & \spacerB\spacerB{}{4.4} & \spacerB\spacerB{}3.3 & \spacerB\spacerB{}5.3 \\ \bottomrule
\end{tabular}
}
\caption{\textbf{Differences to the baseline performance per set-up type} (intermediate-task training (ITT) vs.\ MTL) \textbf{and auxiliary task} (UD, NER, MLM), in percentage points (pp.).
E.g., the results of the intermediate task-training set-up with the UD tasks (UD\sequential{}SID) beat the baseline by 0.3~pp.\ for intent detection and 3.9~pp.\ for slot-filling.
Scores are averaged across the Bavarian test sets and three random seeds.
}
\label{tab:differences-setups}
\end{table*}

We first present the results of the baseline models~(\S\ref{sec:results-baselines}), and then discuss the impact of fine-tuning the model on auxiliary Bavarian NLP tasks~(\S\ref{sec:results-multitask-sequential}). 
We next compare performances across the Bavarian dialects as well as an additional Upper German dialect (Bernese Swiss German) and the standard languages German and English~(\S\ref{sec:results-language-comparison}).
We additionally discuss differences between the Bavarian translations~(\S\ref{sec:results-bavarian-differences}), and lastly analyze the results on other Bavarian SID datasets~(\S\ref{sec:results-other-sid}).

\subsection{Baselines: No Auxiliary Tasks}
\label{sec:results-baselines}

Our baseline experiments with mBERT and \xlmr{} achieve similar scores to the results reported by \citet{van-der-goot-etal-2021-masked} for the overall cross-lingual xSID test sets (Appendix~\ref{sec:appendix-baselines}).
However, these two models perform worse than GBERT and mDeBERTa on the Bavarian test sets (see top of Figure~\ref{fig:overview_all_results_dialects}).
GBERT provides the best slot filling scores (F1: 47.2\%) %
and a slightly higher proportion of fully correctly annotated sentences (15.4\%, mDeBERTA: 15.1\%), while mDeBERTa scores the highest intent detection accuracy (73.5\%). %

For the remaining experiments, we use mDeBERTa as we expect the results of a multilingual model to be more generalizable when applied to other languages/dialects than the ones in our study.

\subsection{Multi-Task Learning and Intermediate-Task Training}
\label{sec:results-multitask-sequential}

Both the choice of joint or sequential setup and the choice of auxiliary tasks influence the results (Table~\ref{tab:differences-setups}).
Generally, the auxiliary tasks are more helpful for slot filling than for intent classification. 
This might be due to them, like slot filling, being on a token level.
We could not include any sentence-level content classification tasks, for lack of datasets (cf.\ \citealp{blaschke-etal-2023-survey}).

Except for the MTL model with all auxiliary tasks, all settings that improve slot filling also help with intent classification, and vice versa.

\paragraph{Joint multi-task vs.\ intermediate-task training}
The \textbf{intermediate-task} setups (i.e., SID as a separate, last task) tend to beat the baseline in terms of both intent detection and slot filling, with gains of between 0.3 and 5.1 percentage points (pp.)\ for intent detection and between 3.8 and 9.0 for slot filling.
The only exception is MLM\sequential{}SID (\mbox{--1.8~pp.}\ for intents, \mbox{--5.8~pp.}\ for slots).%
\footnote{
In the set-ups where the model is first exclusively fine-tuned on MLM, the perplexity on the MLM development set is much higher than otherwise (Table~\ref{tab:aux-dev-results} in Appendix~\S\ref{sec:appendix-auxiliary-dev}), i.e., the auxiliary task was not learned properly.
A possible explanation is that the standard hyperparameters might not have been optimal for MLM, and that the different model parameter updates in a multi-task learning context mitigated this somewhat.
}
We assume that separately fine-tuning the model on SID works well as the SID-related model weights cannot afterwards be modified by other tasks.

The \textbf{joint multi-task} setups 
(where SID is trained simultaneously with the other tasks), 
however, show less clear trends.
Some task combinations have a large negative impact on intent classification (e.g., \mbox{--24.6 pp.}\ for UD\multitask{}SID; \mbox{--15.6 pp.}\ when jointly fine-tuning on all tasks), while others have positive effects (e.g., \mbox{+4.3 pp.}\ for MLM\multitask{}NER\multitask{}SID).
The effect on slot filling
is much more positive, with performance differences ranging from \mbox{--3.2} to +9.5 percentage points.
Here, performance appears to depend more on the choice of auxiliary task:

\paragraph{Auxiliary task choice}
The \textbf{UD} tasks help when they are included as intermediary tasks, but lower the performance in nearly all joint MTL settings. %
This is somewhat similar to the results by \citet{van-der-goot-etal-2021-masked}, who found MTL with target-language UD tasks to mostly lower the intent classification performance but to have a mixed impact on slot filling.

Including \textbf{NER} as an auxiliary task is almost always beneficial for slot filling (and otherwise only has a small impact: \mbox{--0.6 pp.}\ for UD\multitask{}NER\multitask{}SID). We hypothesize that this is due to the high similarity between the two tasks (cf.\ 
\citealp{louvan-magnini-2020-recent}).
It also has a positive effect on the intent classification performance, except in joint setups with UD and SID.

On its own, \textbf{MLM} has a negative effect on both slot filling and intent classification, regardless of whether it is included as a joint or intermediate-task.
When it is, however, used together with other auxiliary tasks, it always improves the slot filling performance and nearly always helps the intent classification performance. 
These findings are somewhat different from the ones by \citet{van-der-goot-etal-2021-masked}, where joint MTL with target-language MLM improves slot filling performance and has mixed effects on intent classification.
It is possible that the MLM dataset in our study is too small to meaningfully serve as data for continued pre-training, and that including more data would have made MLM a more beneficial task.

\begin{figure}[t]
\centering
Intent accuracy\strut\\[-5pt]
\adjustbox{max width=\linewidth}{%
\begin{tikzpicture}[remember picture]
\pgfplotstableread{
Setting    de-by    de-ba    de-st    gsw    de    en
{SID (mBERT)}    61.3    65.7    65.8    48.7    74.8    99.0
{SID (XLM-R)}    55.5    68.9    70.9    47.1    88.9    99.0
{SID (GBERT)}    67.9    69.1    73.8    63.9    82.9    99.2
{SID (mDeBERTa)}    64.6    77.7    78.3    57.5    97.9    99.1
UD\sequential{}SID    64.4    79.4    77.7    59.1    96.4    99.3
UD\multitask{}SID    38.9    54.1    53.8    28.8    92.5    99.2
NER\sequential{}SID    67.7    82.8    79.1    66.2    94.2    99.2
NER\multitask{}SID    66.3    82.8    79.6    65.1    94.9    99.1
MLM\sequential{}SID    62.5    77.8    75.0    58.9    94.4    99.3
MLM\multitask{}SID    61.5    76.9    77.3    56.8    95.3    99.2
UD\sequential{}NER\sequential{}SID    69.9    84.3    80.7    65.4    96.3    99.1
UD\multitask{}NER\sequential{}SID    70.5    83.1    81.5    64.5    95.1    99.3
UD\multitask{}NER\multitask{}SID    54.8    65.4    67.7    46.4    93.0    99.3
MLM\multitask{}UD\sequential{}SID    65.9    79.3    76.6    58.8    94.7    99.1
MLM\sequential{}NER\sequential{}SID    66.7    83.3    80.5    64.3    97.3    99.2
MLM\multitask{}NER\sequential{}SID    69.0    85.8    81.1    69.4    96.0    99.1
MLM\multitask{}NER\multitask{}SID    67.7    84.7    81.2    69.1    95.3    99.4
MLM\multitask{}UD\multitask{}NER\multitask{}SID    49.1    62.1    62.7    42.0    89.8    99.1
}\datatable

\begin{axis}[
   xtick=data,
   xticklabels from table={\datatable}{Setting},
   xticklabel style={rotate=90},
   enlarge x limits={0.07},
   xmax=18.7,
   ymin=15,
   ymax=100,
   enlarge y limits={0.05},
   axis line style={draw opacity=0},
   yticklabel style={font=\small, inner sep=-1pt},
   xticklabel style={font=\small},
   xtick style={draw=none},
   ytick style={draw=none},
   width=10cm,
   height=5cm,
   myline/.style={line width=1.5pt, opacity=0.8},
]
\addplot[myline, draw=colEn] table [y=en, x expr=\coordindex,] {\datatable};
\addplot[myline, draw=colDe] table [y=de, x expr=\coordindex,] {\datatable};
\addplot[myline, draw=colDeBa] table [y=de-ba, x expr=\coordindex,] {\datatable};
\addplot[myline, draw=colDeSt] table [y=de-st, x expr=\coordindex,] {\datatable};
\addplot[myline, draw=colDeBy] table [y=de-by, x expr=\coordindex,] {\datatable};
\addplot[myline, draw=colGsw] table [y=gsw, x expr=\coordindex,] {\datatable};

\node[align=left, text=colEn] at (17.7, 100) { en };
\node[align=left, text=colDe] at (17.7, 90) { de };
\node[align=left, text=colDeBa] at (18.3, 69) { de-ba };
\node[align=left, text=colDeSt] at (18.2, 61) { de-st };
\node[align=left, text=colDeBy] at (18.6, 49) { \munichdata };
\node[align=left, text=colGsw] at (18.0, 39) { gsw };

\foreach \y in {20, 40, 60, 80, 100}
{
    \addplot[mark=none, gray, dotted] coordinates {(-0, \y) (17, \y)};
}
\foreach \x in {0, ..., 17}
{
    \addplot[mark=none, gray, dotted] coordinates {(\x, 15) (\x, 100)};
}

\node at (-0.4, 100) {\small \% };

\coordinate (transparentStart1) at (-0.1, 15);
\coordinate (transparentStart2) at (-0.5, -50);
\end{axis}
\end{tikzpicture}
\begin{tikzpicture}[remember picture, overlay]
    \fill[white, transparency group, opacity=.5] (transparentStart1) rectangle ++(1.25, 3.3);
    \fill[white, transparency group, opacity=.5] (transparentStart2) rectangle ++(1.2, 2.2);
\end{tikzpicture}
}

\vspace{-\baselineskip}
Strict slot F1\strut\\[-5pt]
\adjustbox{max width=\linewidth}{%
\begin{tikzpicture}[remember picture]
\pgfplotstableread{
Setting	de-by	de-ba	de-st	gsw	de	en
{SID (mBERT)}	44.1	43.2	41.3	21.3	68.4	94.1
{SID (XLM-R)}	34.4	35.4	32.1	14.2	73.7	93.8
{SID (GBERT)}	47.1	48.2	46.5	30.0	78.8	93.7
{SID (mDeBERTa)}	43.3	46.7	46.0	20.7	83.1	95.1
UD\sequential{}SID	48.4	50.9	48.5	22.6	80.9	95.1
UD\multitask{}SID	38.6	43.6	44.1	22.8	79.8	95.1
NER\sequential{}SID	53.9	55.3	50.2	30.1	82.7	95.4
NER\multitask{}SID	52.5	55.9	52.9	33.3	82.2	95.0
MLM\sequential{}SID	37.4	40.8	40.5	18.1	78.7	94.8
MLM\multitask{}SID	42.0	45.5	46.3	21.2	82.2	96.1
UD\sequential{}NER\sequential{}SID	53.6	56.4	53.0	30.7	82.3	95.4
UD\multitask{}NER\sequential{}SID	53.2	55.6	52.3	29.3	81.4	94.8
UD\multitask{}NER\multitask{}SID	44.1	45.9	44.1	25.7	78.4	95.3
MLM\multitask{}UD\sequential{}SID	48.4	50.2	48.9	21.8	81.1	94.7
MLM\sequential{}NER\sequential{}SID	51.8	53.4	49.3	29.5	80.0	95.1
MLM\multitask{}NER\sequential{}SID	52.5	56.5	52.1	31.8	82.5	94.4
MLM\multitask{}NER\multitask{}SID	53.7	56.6	54.2	32.7	83.0	95.5
MLM\multitask{}UD\multitask{}NER\multitask{}SID	46.3	50.4	49.3	28.7	81.0	95.6
}\datatable

\begin{axis}[
   xtick=data,
   xticklabels from table={\datatable}{Setting},
   xticklabel style={rotate=90},
   enlarge x limits={0.07},
   xmax=18.7,
   ymin=15,
   ymax=100,
   enlarge y limits={0.05},
   axis line style={draw opacity=0},
   yticklabel style={font=\small, inner sep=-1pt},
   xticklabel style={font=\small},
   xtick style={draw=none},
   ytick style={draw=none},
   width=10cm,
   height=5cm,
   myline/.style={line width=1.5pt, opacity=0.8},
]
\addplot[myline, draw=colEn] table [y=en, x expr=\coordindex,] {\datatable};
\addplot[myline, draw=colDe] table [y=de, x expr=\coordindex,] {\datatable};
\addplot[myline, draw=colDeBa] table [y=de-ba, x expr=\coordindex,] {\datatable};
\addplot[myline, draw=colDeSt] table [y=de-st, x expr=\coordindex,] {\datatable};
\addplot[myline, draw=colDeBy] table [y=de-by, x expr=\coordindex,] {\datatable};
\addplot[myline, draw=colGsw] table [y=gsw, x expr=\coordindex,] {\datatable};

\node[align=left, text=colEn] at (17.7, 96) { en };
\node[align=left, text=colDe] at (17.7, 82) { de };
\node[align=left, text=colDeBa] at (18.3, 57) { de-ba };
\node[align=left, text=colDeSt] at (18.2, 49) { de-st };
\node[align=left, text=colDeBy] at (18.6, 40) { \munichdata };
\node[align=left, text=colGsw] at (18.0, 27) { gsw };

\foreach \y in {20, 40, 60, 80, 100}
{
    \addplot[mark=none, gray, dotted] coordinates {(-0, \y) (17, \y)};
}
\foreach \x in {0, ..., 17}
{
    \addplot[mark=none, gray, dotted] coordinates {(\x, 15) (\x, 100)};
}

\node at (-0.4, 100) {\small \% };

\coordinate (transparentStart1) at (-0.1, 15);
\coordinate (transparentStart2) at (-0.5, -50);
\end{axis}
\end{tikzpicture}
\begin{tikzpicture}[remember picture, overlay]
    \fill[white, transparency group, opacity=.5] (transparentStart1) rectangle ++(1.25, 3.3);
    \fill[white, transparency group, opacity=.5] (transparentStart2) rectangle ++(1.2, 2.2);
\end{tikzpicture}
}
\caption{\textbf{Intent (top) and slot (bottom) scores show similar patterns across experimental set-ups for the test varieties.} 
The scores are averaged across three random seeds (more details are in Appendix~\ref{sec:appendix-details}). 
The pale sections to the left show the scores of baseline models with different PLMs. 
We use lines despite the categorical nature of the x-axis to make the plots easier to compare.}
\label{fig:trends-plots}
\end{figure}
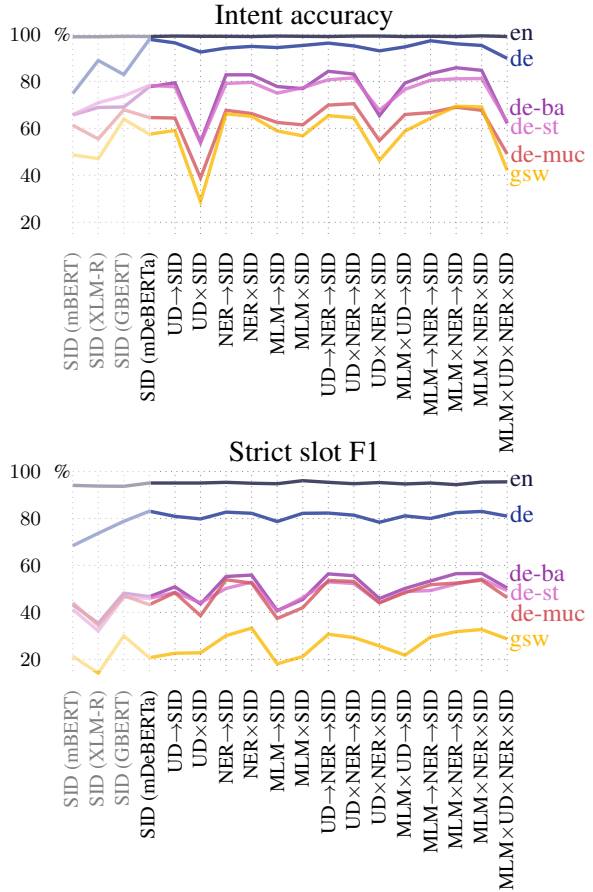

\subsection{Performance Differences Across Languages}
\label{sec:results-language-comparison}

While we previously focused on averages over the three Bavarian dialect datasets, we now compare the performance differences between them, and also analyze the test scores on related languages (Figure~\ref{fig:trends-plots}).
The detailed prediction scores are in Appendix~\ref{sec:appendix-details}, and we summarize the trends below.

\paragraph{Bavarian dialects}
While the scores differ across dialects, the trends across experimental setups are the same: A setup that is beneficial or damaging for the performance on one dialect has a similar effect on the others.
The performance gaps for the multi-task and sequential settings are similar in scale to the gaps of the corresponding baseline. %

The predictions on the Munich Bavarian (\munichdata) test set tend be be worse than for the Upper Bavarian (\deba{}) and South Tyrolean (\dest{}) datasets.
This is especially pronounced for the intent classification results (Figure~\ref{fig:trends-plots}, top).
There, the results on \deba{} and \dest{} are very similar, but the scores on \munichdata{} are between 1.2 and 17.0~pp.\ lower than those on \deba{}.
The slot filling performance is more consistent across dialects (Figure~\ref{fig:trends-plots}, bottom), with score differences of 0.0--5.5~pp.\ between dialect pairs.
Nevertheless, the results on \deba{} tend to be slightly better than for the other dialects.

We discuss differences between the Bavarian test sets in \S\ref{sec:results-bavarian-differences}.

\paragraph{Swiss German}
We additionally consider the performance on Bernese Swiss German, which, like the Bavarian dialects, belongs to the Upper German dialect group.
Performance on Swiss German is always worse than on the Bavarian dialects -- also for the baseline models that were not fine-tuned on Bavarian auxiliary tasks.
This is in line with other SID systems evaluated on the \gsw{} data \cite{aepli-etal-2023-findings} and might be due to the translation being more dissimilar to Standard German than the Bavarian ones (Appendix~\ref{sec:appendix-dataset-distances}).
However, the trends for Swiss German are similar as for the Bavarian dialects: 
Setups that improve or lower SID performance for Bavarian also do so for Swiss German, despite only involving Bavarian auxiliary data.

\paragraph{Standard German} We analyze the performance on Standard German, which is part of mDeBERTa's pretraining dataset.
Performance on Standard German is consistently better than on the Bavarian dialects (intent detection accuracy remains at $\geq89.8$\%, slot filling F1 at $\geq78.7$\%).
Bavarian auxiliary tasks incur performance losses on the Standard German test data across all settings, but the settings that harm performance on Bavarian also have the most deteriorating effect on the predictions for German.

\paragraph{English}
Lastly, we turn to English -- the fine-tuning language. %
The scores %
are barely affected by the auxiliary tasks: Intent detection accuracy remains at $\geq99.1$\% (the same as for the baseline) and slot filling F1 scores at $\geq94.4$\% \mbox{(--0.7 pp.).}

\subsection{Differences Between Bavarian Translations}
\label{sec:results-bavarian-differences}
\begin{table}[t]
\adjustbox{max width=\linewidth}{%
\begin{tabular}{@{}l@{\hspace{5pt}}l@{\hspace{5pt}}l@{\hspace{5pt}}l@{\hspace{5pt}}l@{}}
\dialectname{\munichdata}\\
{streich} & {olle} & {wecka} & & \intent\\[\shortrow]
\gloss{remove.\textsc{imp}} & \gloss{all} & \gloss{alarms} &  \\
\gold{O} & \gold{B--ref.} & \gold{O} & & \gold{alarm/cancel\_alarm} \\
\begin{tabular}[t]{@{}l@{}}\textcolor{darkred}{B--entity}\\\predwrong{\_name}\end{tabular} & \begin{tabular}[t]{@{}l@{}}\textcolor{darkred}{I--rem./}\\\predwrong{todo}\end{tabular} & \predcorrect{O} & & \predwrong{AddToPlaylist}\\
\\
\dialectname{de-ba} \\
{Lösch} & {olle} & {Wegga} & & \intent\\[\shortrow]
\gloss{delete.\textsc{imp}} & \gloss{all} & \gloss{alarms} &  \\
\gold{O} & \gold{B--re.} & \gold{O} & & \gold{alarm/cancel\_alarm}  \\
\predcorrect{O} & \predcorrect{B--ref.} & \predcorrect{O} & & \hspace{-13pt}\predcorrect{alarm/cancel\_alarm} \\
\\
\dialectname{de-st}\\
{tua} & {olle} & {Wecker} & {weck} & \intent \\[\shortrow]
\gloss{do.\textsc{imp}} & \gloss{all} & \gloss{alarms} & \gloss{away} \\
\gold{O} & \gold{B--ref.} & \gold{O} & \gold{O} & \gold{alarm/cancel\_alarm}  \\
\predcorrect{O} & \predwrong{O} & \predcorrect{O} & \predcorrect{O} & \predwrong{alarm/set\_alarm} \\
\end{tabular}}
\caption{\textbf{Translations of ``Delete all alarms'' into Bavarian dialects with \gold{gold-standard} and (\predcorrect{correctly} or \predwrong{incorrectly}) predicted annotations.}
The predictions are by the overall best-performing model, MLM\multitask{}NER\sequential{}SID, with the same random seed.
Abbreviated slots: ref.~=~reference, rem.~=~reminder.
}
\label{tab:delete-all-alarms}
\end{table}

The test sets reflect differences between Bavarian dialects (\S\ref{sec:background-bavarian}) and translation choices.
Table~\ref{tab:delete-all-alarms} shows translations of the English test sentence ``Delete all alarms'', which exhibit both spelling variation (``alarms'' rendered as \textit{wecka, Wegga, Wecker}) and different word choices (\textit{streich} ``remove'', \textit{lösch} ``delete'', and \textit{tua ... weck} ``do ... away'').

Although there is very little morphosyntactic variation between Bavarian dialects, some of the translations exhibit different morphosyntactic structures that reflect different translation choices.
Table~\ref{tab:reminder-get-paper} in Appendix~\ref{sec:appendix-examples} provides an example.

Even small differences between translations can affect the predictions of a SID model.
In both examples, all three translations receive different slot and intent labels by the best-performing model in our experiments -- even though the first two translations in Table~\ref{tab:delete-all-alarms} have an identical structure to the English sentence, which is annotated correctly.

One possible reason for this is that the Munich translation is mostly lower-cased, unlike the other Bavarian translations.
This likely further decreases the subword token overlap with German cognates that might be in the PLM's pretraining data.

\subsection{Additional Bavarian Test Sets}
\label{sec:results-other-sid}
To investigate the robustness of our findings not only across dialects, but also across different datasets from the same area (Upper Bavaria; \deba{}), we use the additional datasets mentioned at the end of \S\ref{sec:data}.
We evaluate the baseline model, the best-performing model (MLM\multitask{}NER\sequential{}SID), and its MTL counterpart (MLM\multitask{}NER\multitask{}SID), which also performs well on the xSID data (Figures \ref{fig:overview_all_results_dialects}~and~\ref{fig:trends-plots}).

All three models perform best on the xSID data (intent accuracy: 77.7\%, slot F1: 46.7\%) and worst on the MASSIVE translations (intents: 55.2\%, slots: 22.1\%), with the naturalistic data in between (intents: 60.8\%, slots: 31.7\%). 
The detailed scores are in Appendix~\ref{sec:appendix-additional-data} (Table~\ref{tab:extra-data-massive-natural}).
The models that were also trained on auxiliary data nearly always improve over the baseline.
The overall best-performing model incurs improvements of \mbox{6.7--7.9 pp.}\ for intent classification and \mbox{9.7--9.9 pp.}\ for slot filling on the additional test sets.
Nevertheless, the magnitudes of the performance gains for each model are slightly different compared to the xSID data.
Thus, while well-performing SID systems are also useful for data from other distributions, the performance patterns are not identical.

\section{Conclusion}
\label{sec:conclusion}

In all of our cross-lingual SID experiments, the performance patterns are similar across dialects, but the actual scores differ.
To allow future research on this kind of variation, we release a new evaluation dataset (\munichdata).
In our experiments, intermediate-task training tends to produce better results than joint multi-task learning.
Additionally, our Bavarian auxiliary tasks (POS tagging and dependency parsing, NER, MLM) were more beneficial for slot filling than intent classification, with NER being the overall most helpful auxiliary task.

\section*{Acknowledgments}

We thank Rob van der Goot for useful discussions regarding MaChAmp, Siyao Peng for early discussions of the topic, Ryan Soh-Eun Shim for giving feedback on an early version of the draft, and the anonymous reviewers for their comments.

This research is supported by European Research Council (ERC) Consolidator Grant DIALECT 101043235.

\section*{Limitations}
\label{sec:limitations}

\paragraph{Data}
The dialect tags should not be taken to reflect all dialect speakers from the respective regions, nor necessarily the most traditional forms of these dialects.
That is, the new \munichdata{} development/\allowbreak{}test set only reflects the language of one young Munich Bavarian speaker (see also~\S\ref{sec:data-statement-limitations}).

\paragraph{Tasks}
Due to lack of data, we could not conduct any experiments with sentence-level auxiliary tasks, and we also could not compare our results to settings with German or even Bavarian SID training data.

We include MLM as one of our auxiliary tasks since it is a common pre-training objective, albeit not the one used for mDeBERTa~v.3 \cite{he2021debertav3}, which instead uses replaced token detection (RTD; \citealp{clark2020electra}).
We use MLM as it is supported by MaChAmp, and selecting a (separate) MLM generator model for RTD would have introduced additional task-specific parameters.

\paragraph{PLMs}
In the paper by \citet{van-der-goot-etal-2021-masked}, the impact of the auxiliary tasks differs for two PLMs. 
Due to computational constraints, we only carried out the (non-baseline) experiments with a single PLM and did not evaluate how robust the results are across PLMs.

\paragraph{Implementation}
We decode the slot predictions with a simple softmax layer.
This might lead to lower slot filling results than decoding the output with conditional random fields to enforce consistent BIO sequences \cite{van-der-goot-etal-2021-masked, van-der-goot-etal-2021-massive}.
We do not assume that changing the output decoder would lead to different trends regarding the effects of MTL and intermediate-task training.

We use MaChAmp's default settings, including the maximum number of epochs (20) to keep feasible computation times. 
In many experiments, the optimal number of epochs was 20 or close to 20.
It is possible that we could have reached better results with a larger number of epochs.
Training the model for longer might have been especially crucial for MLM. 
We hypothesize that this might have increased both the intermediate MLM and the final SID performance of the MLM\sequential{}SID model~(\S\ref{sec:results-multitask-sequential}).

We also use the default settings for all tasks, including MLM.
This leads to the MLM data being split across epochs, leaving only a small portion (70~sentences) being used per epoch.
Disabling this split might have lead to better or more consistent MLM results.

\FloatBarrier
\bibliography{anthology,custom}

\appendix

\section{Dataset Distances}
\label{sec:appendix-dataset-distances}
\begin{table}[t]
\centering
\adjustbox{max width=\textwidth}{%
\setlength{\tabcolsep}{3pt}
\begin{tabular}{@{}l@{}rrrrr}
\toprule
\multicolumn{6}{@{}c@{}}{\textit{Slot similarity (chars), case sensitive}} \\
 & de & de-ba~ & \llap{d}e-mu\rlap{c} & de-st &gsw \\
\midrule
en & \cellcolor[HTML]{8FB9DE}0.51 & \cellcolor[HTML]{8FB8DE}0.51 & \cellcolor[HTML]{82B0DB}0.55 & \cellcolor[HTML]{94BCE0}0.48 & \cellcolor[HTML]{A6C7E5}0.42 \\
de &  & \cellcolor[HTML]{5C99CF}{\color[HTML]{FFFFFF} 0.69} & \cellcolor[HTML]{649ED2}{\color[HTML]{FFFFFF} 0.66} & \cellcolor[HTML]{5293CD}{\color[HTML]{FFFFFF} 0.73} & \cellcolor[HTML]{7BACD9}0.58 \\
de-ba &  &  & \cellcolor[HTML]{478CC9}{\color[HTML]{FFFFFF} 0.77} & \cellcolor[HTML]{5F9BD0}{\color[HTML]{FFFFFF} 0.68} & \cellcolor[HTML]{7FAFDA}0.56 \\
de-muc &  &  &  & \cellcolor[HTML]{639DD1}{\color[HTML]{FFFFFF} 0.67} & \cellcolor[HTML]{8DB8DE}0.51 \\
de-st &  &  &  &  & \cellcolor[HTML]{84B2DB}0.55 \\
\bottomrule
\toprule
\multicolumn{6}{@{}c@{}}{\textit{Slot similarity (chars), case insensitive}} \\
 & de & de-ba~ & \llap{d}e-mu\rlap{c} & de-st &gsw \\
en & \cellcolor[HTML]{87B4DC}0.53 & \cellcolor[HTML]{88B4DC}0.53 & \cellcolor[HTML]{82B0DB}0.55 & \cellcolor[HTML]{8EB8DE}0.51 & \cellcolor[HTML]{9FC3E3}0.45 \\
de &  & \cellcolor[HTML]{5A97CF}{\color[HTML]{FFFFFF} 0.70} & \cellcolor[HTML]{5A98CF}{\color[HTML]{FFFFFF} 0.70} & \cellcolor[HTML]{5192CC}{\color[HTML]{FFFFFF} 0.74} & \cellcolor[HTML]{79ABD8}0.59 \\
de-ba &  &  & \cellcolor[HTML]{3D85C6}{\color[HTML]{FFFFFF} 0.81} & \cellcolor[HTML]{5E9AD0}{\color[HTML]{FFFFFF} 0.69} & \cellcolor[HTML]{7CADD9}0.58 \\
de-muc &  &  &  & \cellcolor[HTML]{5A97CF}{\color[HTML]{FFFFFF} 0.70} & \cellcolor[HTML]{84B2DB}0.54 \\
de-st &  &  &  &  & \cellcolor[HTML]{82B0DA}0.55 \\
\bottomrule
\toprule
\multicolumn{6}{@{}c@{}}{\textit{Sent similarity (words), case sensitive}} \\
 & de & de-ba~ & \llap{d}e-mu\rlap{c} & de-st &gsw \\
en & \cellcolor[HTML]{EEF5FA}0.15 & \cellcolor[HTML]{EBF2F9}0.16 & \cellcolor[HTML]{DBE9F5}0.22 & \cellcolor[HTML]{F1F6FB}0.14 & \cellcolor[HTML]{FFFFFF}0.08 \\
de &  & \cellcolor[HTML]{CDE0F1}0.27 & \cellcolor[HTML]{DFEBF6}0.20 & \cellcolor[HTML]{BED6EC}0.33 & \cellcolor[HTML]{F2F7FB}0.14 \\
de-ba &  &  & \cellcolor[HTML]{9EC2E3}0.45 & \cellcolor[HTML]{C8DCEF}0.29 & \cellcolor[HTML]{F4F8FC}0.13 \\
de-muc &  &  &  & \cellcolor[HTML]{D5E5F3}0.24 & \cellcolor[HTML]{F4F8FC}0.13 \\
de-st &  &  &  &  & \cellcolor[HTML]{F4F8FC}0.13 \\
\bottomrule
\toprule
\multicolumn{6}{@{}c@{}}{\textit{Sent similarity (words), case insensitive}} \\
 & de & de-ba~ & \llap{d}e-mu\rlap{c} & de-st &gsw \\
en & \cellcolor[HTML]{E9F1F9}0.17 & \cellcolor[HTML]{E3EDF7}0.19 & \cellcolor[HTML]{DBE8F5}0.22 & \cellcolor[HTML]{ECF3FA}0.16 & \cellcolor[HTML]{FBFDFE}0.10 \\
de &  & \cellcolor[HTML]{CBDEF0}0.28 & \cellcolor[HTML]{D7E6F4}0.24 & \cellcolor[HTML]{BDD6EC}0.33 & \cellcolor[HTML]{F0F6FB}0.14 \\
de-ba &  &  & \cellcolor[HTML]{8FB9DE}0.50 & \cellcolor[HTML]{C5DBEE}0.30 & \cellcolor[HTML]{F3F7FC}0.13 \\
de-muc &  &  &  & \cellcolor[HTML]{CCDFF1}0.27 & \cellcolor[HTML]{F0F6FB}0.14 \\
de-st &  &  &  &  & \cellcolor[HTML]{F2F7FC}0.13 \\
\bottomrule
\end{tabular}
}
\caption{
\textbf{Mean similarities between slots or sentences corresponding to each other.}
The similarities are calculates as 1~minus the normalized Levenshtein distance.
}
\label{tab:dataset-distances}
\end{table}

We compare how similar the translations are to each other.
For each pair of sentence translations, we calculate the word-level \citet{levenshtein} edit distance.
We also select all words tagged as the same slot type (ignoring the \texttt{B} or \texttt{I} prefixes) and join them with blank spaces.
For corresponding pair of slot values, we calculate the character-level edit distance.
We normalize each distance by dividing it by the length of the longer phrase, and we convert it into a similarity score by subtracting it from~1.

For both similarity levels (sentences and slots) and regardless of whether we consider casing differences,
the two Central Bavarian translations (\deba, \munichdata) are more similar to each other than any of the other pairs (Table~\ref{tab:dataset-distances}).
The Bavarian and Standard German translations are closer to each other than the Swiss German translation.

\section{Data Statement}
\label{sec:appendix-data-statement}

\title{Data Statement for xSID \munichdata{}}
\author{}
\date{}

\subsection{Header}

\begin{itemize}
\item \textit{Dataset Title:} xSID \munichdata{}
\item \textit{Dataset Curator(s):} Xaver Maria Krückl, Verena Blaschke, Barbara Plank
\item \textit{Dataset Version:} 1.0 (expected to be part of xSID~0.7)
\item \textit{Dataset Citation:} Please cite this paper when using this dataset.
\item \textit{Data Statement Authors:} Xaver Maria Krückl, Verena Blaschke, Barbara Plank
\item \textit{Data Statement Version:} 1.0
\item \textit{Data Statement Citation and DOI:} Please cite this paper when referring to the data statement.
\item \textit{Links to versions of this data statement in other languages:} ---
\end{itemize}

\subsection{Executive Summary}

xSID \munichdata{} is a manually annotated (translated) extension of the English xSID development and train set \cite{van-der-goot-etal-2021-masked} into the Bavarian dialect spoken in Munich. The development set contains 300 translated samples and the test set 500. The intents were taken over from the English gold examples whereas the slots were annotated by the translator. 
The translations were made over several weeks. %

\subsection{Curation Rationale}

The purpose of xSID \munichdata{} is to provide further dialectal development and test data in addition to other Bavarian translations. We hope to extend our research on dialectal SID through our data. 

\subsection{Documentation for Source Datasets}

The xSID \munichdata{} development and test set are based on the respective English sets from xSID (\citealp{van-der-goot-etal-2021-masked}; \href{https://github.com/mainlp/xsid/blob/main/LICENSE}{CC BY-SA 4.0}), which in turn are derived in equal parts from two larger datasets, the Snips (\citealp{coucke2018snips}; \href{https://github.com/sonos/nlu-benchmark/blob/master/LICENSE}{CC0 1.0 Universal}) and Facebook (\citealp{schuster-etal-2019-cross-lingual}; \href{https://creativecommons.org/licenses/by-sa/4.0/}{CC-BY-SA license}) datasets.

\subsection{Language Varieties}

xSID \munichdata{} contains data in Munich Bavarian (a Central Bavarian dialect), as spoken by a young speaker. %

\subsection{Language User Demographic}

The original data were created by crowd workers whose demographics are not known.
For the translator, see \textit{Annotator Demographic}.

\subsection{Annotator Demographic}

The translator and annotator is a native speaker of German and Munich Bavarian in his mid-twenties.
He annotated the data while finishing his Master's degree in Computational Linguistics and is one of the authors of this paper.

\subsection{Linguistic Situation and Text Characteristics}

xSID consists of random samples from the English Snips \cite{coucke2018snips} and Facebook \cite{schuster-etal-2019-cross-lingual} datasets, which are compiled from utterances to be used for training digital assistants. Both datasets were mainly crowd-sourced; %
annotations were validated.

\subsection{Preprocessing and Data Formatting}

We directly worked with xSID's English sentences and did not apply any further preprocessing steps. 
Like the rest of xSID, the data set is in the CONLL format.

\subsection{Capture Quality}

Some sentences contain grammatical errors or typos in the original datasets. 
Following xSID's translation guidelines, we retained such errors in the \munichdata{} translations.

\subsection{Limitations}
\label{sec:data-statement-limitations}

The data set is a translation, which probably differs from the way speakers express themselves when not prompted to translate \cite{winkler-etal-2024-slot} or in fluent conversation.

It reflects the language use of a single speaker. 
It does not represent the most traditional form of Munich Bavarian.
Additionally, other speakers might prefer other spellings (since Bavaria has no established orthography).

\subsection{Metadata}

\begin{itemize}
\item \textit{Annotation Guidelines:} Appendices F and G of \citet{van-der-goot-etal-2021-masked}
\item \textit{Annotation Process:}  --- (see this paper)
\item \textit{Dataset Quality Metrics:} ---
\end{itemize}

\subsection{Disclosures and Ethical Review}

There are no conflicts of interest.
This research is supported by European Research Council (ERC) Consolidator Grant DIALECT 101043235.

\subsection{Distribution}

The de-muc split will be included in xSID under the same license, accessible via \url{https://github.com/mainlp/xsid}.

\subsection{Maintenance}

Errors can be reported via GitHub issues or emailing us. Updates to the dataset (and the release history) will be available in the repository.

\subsection{Other}

---

\subsection{Glossary}

---

\smallskip
\subsection*{About this document}

A data statement is a characterization of a dataset that provides context to allow developers and users to better understand how experimental results might generalize, how software might be appropriately deployed, and what biases might be reflected in systems built on the software.

This data statement was written based on the template for the Data Statements Version~3 Schema. The template was prepared by Angelina McMillan-Major and Emily M. Bender and can be found at \url{http://techpolicylab.uw.edu/data-statements}.

\section{Baseline Systems}
\label{sec:appendix-baselines}

\begin{table*}
\centering
\adjustbox{max width=\textwidth}{%
\begin{tabular}{@{}lrrrrrrrrrrrrr}
\toprule
 & \multicolumn{1}{c}{ar} & \multicolumn{1}{c}{da} & \multicolumn{1}{c}{de} & \multicolumn{1}{c}{de-st} & \multicolumn{1}{c}{en} & \multicolumn{1}{c}{id} & \multicolumn{1}{c}{it} & \multicolumn{1}{c}{ja} & \multicolumn{1}{c}{kk} & \multicolumn{1}{c}{nl} & \multicolumn{1}{c}{sr} & \multicolumn{1}{c}{tr} & \multicolumn{1}{c}{zh} \\ \midrule
\multicolumn{10}{l}{\textit{Intents (accuracy, in \%)}} \\
mBERT (vdG) & \cellcolor[HTML]{96D5B6}63.1 & \cellcolor[HTML]{6CC499}87.5 & \cellcolor[HTML]{83CDA9}74.2 & \cellcolor[HTML]{8ED1B0}67.8 & \cellcolor[HTML]{57BB8A}99.7 & \cellcolor[HTML]{78C9A1}80.7 & \cellcolor[HTML]{76C8A0}81.7 & \cellcolor[HTML]{A5DBC1}53.9 & \cellcolor[HTML]{9BD7B9}60.1 & \cellcolor[HTML]{86CEAB}72.3 & \cellcolor[HTML]{80CCA7}75.7 & \cellcolor[HTML]{82CDA8}74.7 & \cellcolor[HTML]{73C79E}83.3 \\
mBERT & \cellcolor[HTML]{8DD1B0}67.9 & \cellcolor[HTML]{71C69C}84.8 & \cellcolor[HTML]{82CDA8}74.8 & \cellcolor[HTML]{91D3B3}65.8 & \cellcolor[HTML]{59BC8B}99.0 & \cellcolor[HTML]{80CCA7}76.0 & \cellcolor[HTML]{7FCCA6}76.3 & \cellcolor[HTML]{A3DABF}55.5 & \cellcolor[HTML]{A0D9BD}56.9 & \cellcolor[HTML]{8AD0AE}69.9 & \cellcolor[HTML]{80CCA7}75.7 & \cellcolor[HTML]{88CFAC}71.3 & \cellcolor[HTML]{71C69C}84.8 \\
XLM-15 (vdG) & \cellcolor[HTML]{92D3B3}65.5 & \cellcolor[HTML]{A1D9BE}56.3 & \cellcolor[HTML]{7BCAA4}78.5 & \cellcolor[HTML]{99D6B8}61.3 & \cellcolor[HTML]{57BB8A}99.7 & \cellcolor[HTML]{C3E7D5}36.4 & \cellcolor[HTML]{AFDFC8}48.0 & \cellcolor[HTML]{BEE5D2}39.1 & \cellcolor[HTML]{CEEBDD}29.9 & \cellcolor[HTML]{B4E1CB}45.4 & \cellcolor[HTML]{BAE4CF}41.4 & \cellcolor[HTML]{8ED2B1}67.3 & \cellcolor[HTML]{7BCAA3}78.8 \\
XLM-R & \cellcolor[HTML]{7CCAA4}78.1 & \cellcolor[HTML]{5FBF90}95.3 & \cellcolor[HTML]{6AC397}88.9 & \cellcolor[HTML]{88CFAD}70.9 & \cellcolor[HTML]{59BC8B}99.0 & \cellcolor[HTML]{5FBF90}95.3 & \cellcolor[HTML]{78C9A1}80.5 & \cellcolor[HTML]{A4DBC0}54.5 & \cellcolor[HTML]{80CCA7}75.8 & \cellcolor[HTML]{72C69D}84.3 & \cellcolor[HTML]{74C79F}82.7 & \cellcolor[HTML]{61BF91}94.1 & \cellcolor[HTML]{5EBE8F}96.0 \\
GBERT & \cellcolor[HTML]{D3EDE0}27.2 & \cellcolor[HTML]{98D5B7}61.9 & \cellcolor[HTML]{74C79E}82.9 & \cellcolor[HTML]{83CDA9}73.8 & \cellcolor[HTML]{58BC8B}99.2 & \cellcolor[HTML]{B1E0C9}46.9 & \cellcolor[HTML]{A8DCC3}52.3 & \cellcolor[HTML]{F7FCFA}5.6 & \cellcolor[HTML]{C5E8D7}34.9 & \cellcolor[HTML]{9CD7BA}59.1 & \cellcolor[HTML]{B3E1CA}45.7 & \cellcolor[HTML]{B2E0C9}46.6 & \cellcolor[HTML]{D8F0E4}23.8 \\
mDeBERTa & \cellcolor[HTML]{6DC49A}86.9 & \cellcolor[HTML]{5DBE8E}96.5 & \cellcolor[HTML]{5BBD8D}97.9 & \cellcolor[HTML]{7CCAA4}78.3 & \cellcolor[HTML]{59BC8B}99.1 & \cellcolor[HTML]{5DBE8F}96.3 & \cellcolor[HTML]{5BBD8D}97.4 & \cellcolor[HTML]{7ACAA3}79.2 & \cellcolor[HTML]{68C296}89.9 & \cellcolor[HTML]{5DBE8E}96.5 & \cellcolor[HTML]{69C397}89.1 & \cellcolor[HTML]{5CBD8D}97.2 & \cellcolor[HTML]{5CBD8E}96.9 \\
\midrule
\multicolumn{10}{l}{\textit{Slots (strict slot F1, in \%)}} \\
mBERT (vdG) & \cellcolor[HTML]{B3E1CA}45.8 & \cellcolor[HTML]{83CDA9}73.9 & \cellcolor[HTML]{C9E9D9}33.0 & \cellcolor[HTML]{AEDFC7}48.5 & \cellcolor[HTML]{5BBD8D}97.6 & \cellcolor[HTML]{88CFAC}71.1 & \cellcolor[HTML]{81CCA8}75.0 & \cellcolor[HTML]{9BD7BA}59.9 & \cellcolor[HTML]{AEDFC7}48.5 & \cellcolor[HTML]{78C9A1}80.4 & \cellcolor[HTML]{8ED2B1}67.4 & \cellcolor[HTML]{A2DABE}55.7 & \cellcolor[HTML]{85CEAA}72.9 \\
mBERT & \cellcolor[HTML]{A8DCC2}52.4 & \cellcolor[HTML]{89D0AD}70.3 & \cellcolor[HTML]{8DD1AF}68.4 & \cellcolor[HTML]{BBE4D0}41.3 & \cellcolor[HTML]{61BF91}94.1 & \cellcolor[HTML]{94D4B5}63.8 & \cellcolor[HTML]{8AD0AE}69.9 & \cellcolor[HTML]{BEE5D2}39.4 & \cellcolor[HTML]{CAEADA}32.2 & \cellcolor[HTML]{8AD0AD}70.1 & \cellcolor[HTML]{A3DABF}55.0 & \cellcolor[HTML]{C9E9D9}32.9 & \cellcolor[HTML]{AFDFC8}48.0 \\
XLM-15 (vdG) & \cellcolor[HTML]{ADDEC6}49.1 & \cellcolor[HTML]{D4EEE1}26.3 & \cellcolor[HTML]{C8E9D9}33.3 & \cellcolor[HTML]{BEE5D2}39.4 & \cellcolor[HTML]{5CBD8E}97.0 & \cellcolor[HTML]{E7F6EF}14.9 & \cellcolor[HTML]{D2EDE0}27.3 & \cellcolor[HTML]{C8E9D9}33.4 & \cellcolor[HTML]{EEF8F3}10.9 & \cellcolor[HTML]{CCEBDC}30.9 & \cellcolor[HTML]{E6F5EE}15.9 & \cellcolor[HTML]{B3E1CB}45.5 & \cellcolor[HTML]{9FD8BC}57.6 \\
XLM-R & \cellcolor[HTML]{97D5B7}62.3 & \cellcolor[HTML]{77C8A1}80.9 & \cellcolor[HTML]{84CDA9}73.7 & \cellcolor[HTML]{CAEADA}32.1 & \cellcolor[HTML]{62C091}93.8 & \cellcolor[HTML]{7FCBA6}76.6 & \cellcolor[HTML]{80CCA7}75.6 & \cellcolor[HTML]{AADDC4}51.0 & \cellcolor[HTML]{B4E1CB}45.2 & \cellcolor[HTML]{75C89F}82.2 & \cellcolor[HTML]{94D4B5}63.9 & \cellcolor[HTML]{A7DCC2}52.9 & \cellcolor[HTML]{8FD2B1}66.8 \\
GBERT & \cellcolor[HTML]{DFF2E9}19.7 & \cellcolor[HTML]{C1E6D4}37.3 & \cellcolor[HTML]{7BCAA3}78.8 & \cellcolor[HTML]{B2E0C9}46.5 & \cellcolor[HTML]{62C092}93.7 & \cellcolor[HTML]{E3F4EC}17.2 & \cellcolor[HTML]{D1EDDF}28.0 & \cellcolor[HTML]{FFFFFF}0.7 & \cellcolor[HTML]{F8FCFA}5.4 & \cellcolor[HTML]{B5E1CC}44.4 & \cellcolor[HTML]{E1F3EA}18.5 & \cellcolor[HTML]{F3FAF7}8.3 & \cellcolor[HTML]{E8F6EF}14.5 \\
mDeBERTa & \cellcolor[HTML]{88CFAC}71.1 & \cellcolor[HTML]{79C9A2}79.7 & \cellcolor[HTML]{74C79E}83.1 & \cellcolor[HTML]{B3E0CA}46.0 & \cellcolor[HTML]{5FBF90}95.1 & \cellcolor[HTML]{7CCAA4}78.3 & \cellcolor[HTML]{74C79E}83.1 & \cellcolor[HTML]{ACDEC5}49.8 & \cellcolor[HTML]{A8DCC2}52.4 & \cellcolor[HTML]{6EC49A}86.6 & \cellcolor[HTML]{86CEAB}72.1 & \cellcolor[HTML]{9ED8BB}58.3 & \cellcolor[HTML]{82CDA8}74.7 \\
\midrule
\multicolumn{10}{l}{\textit{Fully correct (in \%)}} \\
mBERT & \cellcolor[HTML]{E1F3EA}18.5 & \cellcolor[HTML]{B5E1CC}44.5 & \cellcolor[HTML]{C6E8D7}34.6 & \cellcolor[HTML]{F1F9F5}9.5 & \cellcolor[HTML]{6BC398}88.3 & \cellcolor[HTML]{CBEADB}31.6 & \cellcolor[HTML]{C1E6D4}37.7 & \cellcolor[HTML]{DEF2E8}20.3 & \cellcolor[HTML]{F2FAF6}8.5 & \cellcolor[HTML]{C2E6D4}37.1 & \cellcolor[HTML]{D7EFE3}24.6 & \cellcolor[HTML]{ECF7F2}12.4 & \cellcolor[HTML]{E6F5EE}15.9 \\
XLM-R & \cellcolor[HTML]{D0ECDE}28.8 & \cellcolor[HTML]{93D4B4}64.4 & \cellcolor[HTML]{ADDEC6}49.3 & \cellcolor[HTML]{F5FBF8}6.9 & \cellcolor[HTML]{6BC398}88.5 & \cellcolor[HTML]{A2DABE}56.0 & \cellcolor[HTML]{B0DFC8}47.4 & \cellcolor[HTML]{D5EEE2}25.9 & \cellcolor[HTML]{E4F4EC}16.9 & \cellcolor[HTML]{9FD8BC}57.7 & \cellcolor[HTML]{C0E6D3}38.0 & \cellcolor[HTML]{C6E8D7}34.6 & \cellcolor[HTML]{B2E0C9}46.4 \\
GBERT & \cellcolor[HTML]{F8FDFB}4.9 & \cellcolor[HTML]{ECF7F2}12.4 & \cellcolor[HTML]{A6DBC1}53.7 & \cellcolor[HTML]{E8F6EF}14.7 & \cellcolor[HTML]{6CC499}87.5 & \cellcolor[HTML]{FDFFFE}1.9 & \cellcolor[HTML]{F9FDFB}4.5 & \cellcolor[HTML]{FFFFFF}0.9 & \cellcolor[HTML]{FDFFFE}1.9 & \cellcolor[HTML]{EBF7F1}12.8 & \cellcolor[HTML]{FAFDFC}3.9 & \cellcolor[HTML]{FDFEFE}2.3 & \cellcolor[HTML]{FBFEFC}3.3 \\
mDeBERTa & \cellcolor[HTML]{B6E2CC}44.1 & \cellcolor[HTML]{98D6B7}61.7 & \cellcolor[HTML]{91D3B2}66.1 & \cellcolor[HTML]{E6F5EE}15.9 & \cellcolor[HTML]{68C296}90.0 & \cellcolor[HTML]{9DD8BB}58.8 & \cellcolor[HTML]{96D5B6}63.1 & \cellcolor[HTML]{BCE4D1}40.4 & \cellcolor[HTML]{D7EFE3}24.6 & \cellcolor[HTML]{85CEAA}72.7 & \cellcolor[HTML]{B2E0CA}46.3 & \cellcolor[HTML]{C4E8D6}35.6 & \cellcolor[HTML]{9FD8BC}57.7 \\ \bottomrule
\end{tabular}
}
\caption{\textbf{Scores of our baselines on xSID's original test language selection.} We also include scores by \citet{van-der-goot-etal-2021-masked} for comparison (=~vdG).
XLM-15 refers to \href{https://huggingface.co/FacebookAI/xlm-mlm-tlm-xnli15-1024}{xlm-mlm-tlm-xnli15-1024} \cite{xlm}.}
\label{tab:baseline-results}
\end{table*}

Table~\ref{tab:baseline-results} shows the results of our baseline systems (no auxiliary tasks) and the baseline systems by \citet{van-der-goot-etal-2021-masked} on all languages that were in the original xSID release.
Note that we use XLM-R while \citet{van-der-goot-etal-2021-masked} use XLM-15.

\section{Detailed Results}
\label{sec:appendix-details}

\begin{table*}
\centering
\begin{tabular}{@{}lrrrrrr}
\toprule
 & \multicolumn{1}{c}{\munichdata} & \multicolumn{1}{c}{de-ba} & \multicolumn{1}{c}{de-st} & \multicolumn{1}{c}{gsw} & \multicolumn{1}{c}{de} & \multicolumn{1}{c}{en} \\ \midrule
SID (mBERT) & \cellcolor[HTML]{99D6B8}61.3\textsubscript{1.2} & \cellcolor[HTML]{91D3B3}65.7\textsubscript{2.4} & \cellcolor[HTML]{91D3B3}65.8\textsubscript{2.0} & \cellcolor[HTML]{AEDEC7}48.7\textsubscript{2.2} & \cellcolor[HTML]{82CDA8}74.8\textsubscript{1.7} & \cellcolor[HTML]{59BC8C}99.0\textsubscript{0.2} \\
SID (XLM-R) & \cellcolor[HTML]{A2DABF}55.5\textsubscript{2.6} & \cellcolor[HTML]{8CD1AF}68.9\textsubscript{0.7} & \cellcolor[HTML]{88CFAD}70.9\textsubscript{1.2} & \cellcolor[HTML]{B0DFC8}47.1\textsubscript{2.6} & \cellcolor[HTML]{6AC397}88.9\textsubscript{1.7} & \cellcolor[HTML]{59BC8C}99.0\textsubscript{0.2} \\
SID (GBERT) & \cellcolor[HTML]{8DD1B0}67.9\textsubscript{2.7} & \cellcolor[HTML]{8BD1AF}69.1\textsubscript{0.5} & \cellcolor[HTML]{84CDA9}73.8\textsubscript{1.0} & \cellcolor[HTML]{94D4B5}63.9\textsubscript{1.5} & \cellcolor[HTML]{74C79F}82.9\textsubscript{1.3} & \cellcolor[HTML]{59BC8B}99.2\textsubscript{0.0} \\
SID (mDeBERTa) & \cellcolor[HTML]{93D4B4}64.6\textsubscript{3.1} & \cellcolor[HTML]{7DCBA5}77.7\textsubscript{0.7} & \cellcolor[HTML]{7CCAA4}78.3\textsubscript{0.8} & \cellcolor[HTML]{9FD8BC}57.5\textsubscript{2.7} & \cellcolor[HTML]{5BBD8D}97.9\textsubscript{0.4} & \cellcolor[HTML]{59BC8C}99.1\textsubscript{0.1} \\
UD\sequential{}SID & \cellcolor[HTML]{93D4B4}64.4\textsubscript{4.0} & \cellcolor[HTML]{7ACAA3}79.4\textsubscript{2.3} & \cellcolor[HTML]{7DCBA5}77.7\textsubscript{2.4} & \cellcolor[HTML]{9CD7BA}59.1\textsubscript{4.6} & \cellcolor[HTML]{5EBE8F}96.4\textsubscript{1.0} & \cellcolor[HTML]{59BC8B}99.3\textsubscript{0.2} \\
UD\multitask{}SID & \cellcolor[HTML]{BEE5D2}38.9\textsubscript{3.2} & \cellcolor[HTML]{A5DBC0}54.1\textsubscript{2.8} & \cellcolor[HTML]{A5DBC1}53.8\textsubscript{2.5} & \cellcolor[HTML]{CFECDE}28.8\textsubscript{1.4} & \cellcolor[HTML]{64C193}92.5\textsubscript{1.7} & \cellcolor[HTML]{59BC8B}99.2\textsubscript{0.2} \\
NER\sequential{}SID & \cellcolor[HTML]{8ED1B0}67.7\textsubscript{0.5} & \cellcolor[HTML]{74C79F}82.8\textsubscript{3.5} & \cellcolor[HTML]{7BCAA3}79.1\textsubscript{1.3} & \cellcolor[HTML]{90D2B2}66.2\textsubscript{3.2} & \cellcolor[HTML]{61BF91}94.2\textsubscript{0.7} & \cellcolor[HTML]{59BC8B}99.2\textsubscript{0.2} \\
NER\multitask{}SID & \cellcolor[HTML]{90D2B2}66.3\textsubscript{1.6} & \cellcolor[HTML]{74C79F}82.8\textsubscript{1.8} & \cellcolor[HTML]{7AC9A2}79.6\textsubscript{1.0} & \cellcolor[HTML]{92D3B3}65.1\textsubscript{1.1} & \cellcolor[HTML]{60BF90}94.9\textsubscript{1.2} & \cellcolor[HTML]{59BC8C}99.1\textsubscript{0.1} \\
MLM\sequential{}SID & \cellcolor[HTML]{96D5B6}62.5\textsubscript{2.2} & \cellcolor[HTML]{7DCBA4}77.8\textsubscript{2.9} & \cellcolor[HTML]{81CCA8}75.0\textsubscript{1.8} & \cellcolor[HTML]{9DD7BB}58.9\textsubscript{5.2} & \cellcolor[HTML]{61BF91}94.4\textsubscript{2.4} & \cellcolor[HTML]{59BC8B}99.3\textsubscript{0.2} \\
MLM\multitask{}SID & \cellcolor[HTML]{98D6B8}61.5\textsubscript{2.1} & \cellcolor[HTML]{7ECBA6}76.9\textsubscript{2.1} & \cellcolor[HTML]{7ECBA5}77.3\textsubscript{0.5} & \cellcolor[HTML]{A0D9BD}56.8\textsubscript{3.4} & \cellcolor[HTML]{5FBF90}95.3\textsubscript{1.3} & \cellcolor[HTML]{59BC8B}99.2\textsubscript{0.2} \\
UD\sequential{}NER\sequential{}SID & \cellcolor[HTML]{8AD0AE}69.9\textsubscript{1.0} & \cellcolor[HTML]{72C69D}84.3\textsubscript{2.9} & \cellcolor[HTML]{78C9A1}80.7\textsubscript{1.5} & \cellcolor[HTML]{92D3B3}65.4\textsubscript{1.8} & \cellcolor[HTML]{5EBE8F}96.3\textsubscript{0.7} & \cellcolor[HTML]{59BC8C}99.1\textsubscript{0.1} \\
UD\multitask{}NER\sequential{}SID & \cellcolor[HTML]{89D0AD}70.5\textsubscript{1.8} & \cellcolor[HTML]{74C79E}83.1\textsubscript{1.7} & \cellcolor[HTML]{77C8A0}81.5\textsubscript{1.2} & \cellcolor[HTML]{93D4B4}64.5\textsubscript{3.6} & \cellcolor[HTML]{60BF90}95.1\textsubscript{2.7} & \cellcolor[HTML]{59BC8B}99.3\textsubscript{0.2} \\
UD\multitask{}NER\multitask{}SID & \cellcolor[HTML]{A3DABF}54.8\textsubscript{2.4} & \cellcolor[HTML]{92D3B3}65.4\textsubscript{3.5} & \cellcolor[HTML]{8ED1B0}67.7\textsubscript{1.1} & \cellcolor[HTML]{B2E0C9}46.4\textsubscript{3.1} & \cellcolor[HTML]{63C093}93.0\textsubscript{0.9} & \cellcolor[HTML]{59BC8B}99.3\textsubscript{0.1} \\
MLM\multitask{}UD\sequential{}SID & \cellcolor[HTML]{91D3B2}65.9\textsubscript{1.0} & \cellcolor[HTML]{7ACAA3}79.3\textsubscript{1.4} & \cellcolor[HTML]{7FCBA6}76.6\textsubscript{2.5} & \cellcolor[HTML]{9DD8BB}58.8\textsubscript{1.0} & \cellcolor[HTML]{60BF91}94.7\textsubscript{1.5} & \cellcolor[HTML]{59BC8C}99.1\textsubscript{0.2} \\
MLM\sequential{}NER\sequential{}SID & \cellcolor[HTML]{8FD2B1}66.7\textsubscript{1.0} & \cellcolor[HTML]{74C79E}83.3\textsubscript{1.6} & \cellcolor[HTML]{78C9A1}80.5\textsubscript{1.1} & \cellcolor[HTML]{93D4B4}64.3\textsubscript{1.9} & \cellcolor[HTML]{5CBD8E}97.3\textsubscript{0.5} & \cellcolor[HTML]{59BC8B}99.2\textsubscript{0.2} \\
MLM\multitask{}NER\sequential{}SID & \cellcolor[HTML]{8CD1AF}69.0\textsubscript{1.7} & \cellcolor[HTML]{6FC59B}85.8\textsubscript{1.3} & \cellcolor[HTML]{77C8A1}81.1\textsubscript{0.7} & \cellcolor[HTML]{8BD0AE}69.4\textsubscript{1.7} & \cellcolor[HTML]{5EBE8F}96.0\textsubscript{1.1} & \cellcolor[HTML]{59BC8C}99.1\textsubscript{0.1} \\
MLM\multitask{}NER\multitask{}SID & \cellcolor[HTML]{8ED1B0}67.7\textsubscript{1.0} & \cellcolor[HTML]{71C69C}84.7\textsubscript{0.5} & \cellcolor[HTML]{77C8A0}81.2\textsubscript{2.6} & \cellcolor[HTML]{8BD1AF}69.1\textsubscript{3.3} & \cellcolor[HTML]{5FBF90}95.3\textsubscript{1.0} & \cellcolor[HTML]{59BC8B}99.4\textsubscript{0.2} \\
MLM\multitask{}UD\multitask{}NER\multitask{}SID & \cellcolor[HTML]{ADDEC6}49.1\textsubscript{5.8} & \cellcolor[HTML]{97D5B7}62.1\textsubscript{4.6} & \cellcolor[HTML]{96D5B6}62.7\textsubscript{3.6} & \cellcolor[HTML]{B9E3CE}42.0\textsubscript{8.3} & \cellcolor[HTML]{69C296}89.8\textsubscript{0.9} & \cellcolor[HTML]{59BC8C}99.1\textsubscript{0.2} \\
\bottomrule
\end{tabular}
\caption{\textbf{Intent classification results in the three Bavarian dialects, Swiss German, German, and English.}
We show mean scores (accuracy, in~\%) over three random seeds, with standard deviations in subscripts.
}
\label{tab:results-detailed-intents}
\end{table*}

\begin{table*}
\centering
\begin{tabular}{@{}lrrrrrr}
\toprule
 & \multicolumn{1}{c}{\munichdata} & \multicolumn{1}{c}{de-ba} & \multicolumn{1}{c}{de-st} & \multicolumn{1}{c}{gsw} &  \multicolumn{1}{c}{de} & \multicolumn{1}{c}{en} \\ \midrule
SID (mBERT) & \cellcolor[HTML]{B5E2CC}44.1\textsubscript{1.1} & \cellcolor[HTML]{B7E2CD}43.2\textsubscript{1.1} & \cellcolor[HTML]{BAE3CF}41.3\textsubscript{1.0} & \cellcolor[HTML]{DCF1E7}21.3\textsubscript{1.3} & \cellcolor[HTML]{8DD1AF}68.4\textsubscript{1.2} & \cellcolor[HTML]{61C091}94.1\textsubscript{0.3} \\
SID (XLM-R) & \cellcolor[HTML]{C6E8D7}34.4\textsubscript{1.6} & \cellcolor[HTML]{C4E7D6}35.4\textsubscript{0.7} & \cellcolor[HTML]{CAEADA}32.1\textsubscript{0.7} & \cellcolor[HTML]{E8F6EF}14.2\textsubscript{0.6} & \cellcolor[HTML]{84CDA9}73.7\textsubscript{1.3} & \cellcolor[HTML]{62C092}93.8\textsubscript{0.5} \\
SID (GBERT) & \cellcolor[HTML]{B0DFC8}47.1\textsubscript{1.3} & \cellcolor[HTML]{AFDFC7}48.2\textsubscript{0.7} & \cellcolor[HTML]{B1E0C9}46.5\textsubscript{2.6} & \cellcolor[HTML]{CDEBDC}30.0\textsubscript{0.9} & \cellcolor[HTML]{7BCAA3}78.8\textsubscript{0.8} & \cellcolor[HTML]{62C092}93.7\textsubscript{0.4} \\
SID (mDeBERTa) & \cellcolor[HTML]{B7E2CD}43.3\textsubscript{0.9} & \cellcolor[HTML]{B1E0C9}46.7\textsubscript{2.0} & \cellcolor[HTML]{B2E0CA}46.0\textsubscript{1.0} & \cellcolor[HTML]{DDF1E7}20.7\textsubscript{2.5} & \cellcolor[HTML]{74C79E}83.1\textsubscript{0.7} & \cellcolor[HTML]{60BF90}95.1\textsubscript{0.2} \\
UD\sequential{}SID & \cellcolor[HTML]{AEDFC7}48.4\textsubscript{2.5} & \cellcolor[HTML]{AADDC4}50.9\textsubscript{0.2} & \cellcolor[HTML]{AEDFC7}48.5\textsubscript{2.2} & \cellcolor[HTML]{DAF0E5}22.6\textsubscript{0.3} & \cellcolor[HTML]{78C8A1}80.9\textsubscript{1.5} & \cellcolor[HTML]{60BF90}95.1\textsubscript{0.1} \\
UD\multitask{}SID & \cellcolor[HTML]{BFE5D2}38.6\textsubscript{2.9} & \cellcolor[HTML]{B6E2CC}43.6\textsubscript{4.2} & \cellcolor[HTML]{B5E2CC}44.1\textsubscript{4.0} & \cellcolor[HTML]{D9F0E5}22.8\textsubscript{3.4} & \cellcolor[HTML]{79C9A2}79.8\textsubscript{1.2} & \cellcolor[HTML]{60BF90}95.1\textsubscript{0.2} \\
NER\sequential{}SID & \cellcolor[HTML]{A5DBC0}53.9\textsubscript{0.5} & \cellcolor[HTML]{A3DABF}55.3\textsubscript{2.3} & \cellcolor[HTML]{ABDDC5}50.2\textsubscript{1.4} & \cellcolor[HTML]{CDEBDC}30.1\textsubscript{1.4} & \cellcolor[HTML]{75C79F}82.7\textsubscript{0.6} & \cellcolor[HTML]{5FBF90}95.4\textsubscript{0.3} \\
NER\multitask{}SID & \cellcolor[HTML]{A7DCC2}52.5\textsubscript{1.8} & \cellcolor[HTML]{A2D9BE}55.9\textsubscript{1.1} & \cellcolor[HTML]{A7DCC2}52.9\textsubscript{0.9} & \cellcolor[HTML]{C8E9D9}33.3\textsubscript{0.1} & \cellcolor[HTML]{75C89F}82.2\textsubscript{1.1} & \cellcolor[HTML]{60BF90}95.0\textsubscript{0.2} \\
MLM\sequential{}SID & \cellcolor[HTML]{C1E6D4}37.4\textsubscript{2.9} & \cellcolor[HTML]{BBE4D0}40.8\textsubscript{4.1} & \cellcolor[HTML]{BBE4D0}40.5\textsubscript{3.4} & \cellcolor[HTML]{E1F3EA}18.1\textsubscript{2.0} & \cellcolor[HTML]{7BCAA3}78.7\textsubscript{1.6} & \cellcolor[HTML]{60BF91}94.8\textsubscript{0.6} \\
MLM\multitask{}SID & \cellcolor[HTML]{B9E3CE}42.0\textsubscript{1.4} & \cellcolor[HTML]{B3E1CA}45.5\textsubscript{2.0} & \cellcolor[HTML]{B2E0C9}46.3\textsubscript{1.6} & \cellcolor[HTML]{DCF1E7}21.2\textsubscript{1.8} & \cellcolor[HTML]{75C89F}82.2\textsubscript{0.7} & \cellcolor[HTML]{5EBE8F}96.1\textsubscript{0.2} \\
UD\sequential{}NER\sequential{}SID & \cellcolor[HTML]{A5DBC1}53.6\textsubscript{1.9} & \cellcolor[HTML]{A1D9BE}56.4\textsubscript{3.6} & \cellcolor[HTML]{A6DBC1}53.0\textsubscript{3.2} & \cellcolor[HTML]{CCEBDC}30.7\textsubscript{2.4} & \cellcolor[HTML]{75C89F}82.3\textsubscript{1.9} & \cellcolor[HTML]{5FBF90}95.4\textsubscript{0.5} \\
UD\multitask{}NER\sequential{}SID & \cellcolor[HTML]{A6DBC1}53.2\textsubscript{1.6} & \cellcolor[HTML]{A2DABE}55.6\textsubscript{1.7} & \cellcolor[HTML]{A8DCC2}52.3\textsubscript{0.5} & \cellcolor[HTML]{CEECDD}29.3\textsubscript{0.4} & \cellcolor[HTML]{77C8A0}81.4\textsubscript{1.3} & \cellcolor[HTML]{60BF91}94.8\textsubscript{0.7} \\
UD\multitask{}NER\multitask{}SID & \cellcolor[HTML]{B5E2CC}44.1\textsubscript{2.6} & \cellcolor[HTML]{B2E0CA}45.9\textsubscript{5.1} & \cellcolor[HTML]{B5E2CC}44.1\textsubscript{5.3} & \cellcolor[HTML]{D4EEE1}25.7\textsubscript{5.0} & \cellcolor[HTML]{7CCAA4}78.4\textsubscript{2.4} & \cellcolor[HTML]{5FBF90}95.3\textsubscript{0.4} \\
MLM\multitask{}UD\sequential{}SID & \cellcolor[HTML]{AEDFC7}48.4\textsubscript{2.6} & \cellcolor[HTML]{ABDDC5}50.2\textsubscript{3.3} & \cellcolor[HTML]{ADDEC6}48.9\textsubscript{1.8} & \cellcolor[HTML]{DBF1E6}21.8\textsubscript{2.2} & \cellcolor[HTML]{77C8A1}81.1\textsubscript{0.9} & \cellcolor[HTML]{60BF91}94.7\textsubscript{0.3} \\
MLM\sequential{}NER\sequential{}SID & \cellcolor[HTML]{A8DCC3}51.8\textsubscript{1.7} & \cellcolor[HTML]{A6DBC1}53.4\textsubscript{0.9} & \cellcolor[HTML]{ADDEC6}49.3\textsubscript{1.4} & \cellcolor[HTML]{CEEBDD}29.5\textsubscript{0.9} & \cellcolor[HTML]{79C9A2}80.0\textsubscript{0.3} & \cellcolor[HTML]{60BF90}95.1\textsubscript{0.3} \\
MLM\multitask{}NER\sequential{}SID & \cellcolor[HTML]{A7DCC2}52.5\textsubscript{1.1} & \cellcolor[HTML]{A1D9BD}56.5\textsubscript{0.9} & \cellcolor[HTML]{A8DCC3}52.1\textsubscript{1.6} & \cellcolor[HTML]{CAEADA}31.8\textsubscript{1.9} & \cellcolor[HTML]{75C79F}82.5\textsubscript{0.4} & \cellcolor[HTML]{61BF91}94.4\textsubscript{0.8} \\ 
MLM\multitask{}NER\multitask{}SID & \cellcolor[HTML]{A5DBC1}53.7\textsubscript{1.1} & \cellcolor[HTML]{A0D9BD}56.6\textsubscript{0.3} & \cellcolor[HTML]{A4DBC0}54.2\textsubscript{1.5} & \cellcolor[HTML]{C9E9D9}32.7\textsubscript{0.9} & \cellcolor[HTML]{74C79E}83.0\textsubscript{0.9} & \cellcolor[HTML]{5FBF90}95.5\textsubscript{0.3} \\
MLM\multitask{}UD\multitask{}NER\multitask{}SID & \cellcolor[HTML]{B2E0C9}46.3\textsubscript{1.2} & \cellcolor[HTML]{ABDDC5}50.4\textsubscript{2.3} & \cellcolor[HTML]{ADDEC6}49.3\textsubscript{1.1} & \cellcolor[HTML]{CFECDE}28.7\textsubscript{0.8} & \cellcolor[HTML]{77C8A1}81.0\textsubscript{0.7} & \cellcolor[HTML]{5FBE90}95.6\textsubscript{0.4} \\
\bottomrule
\end{tabular}
\caption{\textbf{Slots classification results in the three Bavarian dialects, Swiss German, German, and English.}
We show mean scores (strict slot F1, in~\%) over three random seeds, with standard deviations in subscripts.
}
\label{tab:results-detailed-slots}
\end{table*}

\begin{table*}
\centering
\begin{tabular}{@{}lrrrrrr}
\toprule
 & \multicolumn{1}{c}{\munichdata} & \multicolumn{1}{c}{de-ba} & \multicolumn{1}{c}{de-st} & \multicolumn{1}{c}{gsw} & \multicolumn{1}{c}{de} & \multicolumn{1}{c}{en} \\ \midrule
SID (mBERT) & \cellcolor[HTML]{EDF8F3}11.0\textsubscript{0.2} & \cellcolor[HTML]{E9F6F0}13.4\textsubscript{0.3} & \cellcolor[HTML]{F0F9F4}9.5\textsubscript{0.2} & \cellcolor[HTML]{FAFDFC}3.0\textsubscript{0.3} & \cellcolor[HTML]{C5E8D7}34.6\textsubscript{1.6} & \cellcolor[HTML]{6BC398}88.3\textsubscript{0.2} \\
SID (XLM-R) & \cellcolor[HTML]{F5FBF8}6.3\textsubscript{1.1} & \cellcolor[HTML]{EDF8F2}11.3\textsubscript{1.2} & \cellcolor[HTML]{F4FBF7}6.9\textsubscript{0.8} & \cellcolor[HTML]{FDFEFE}1.6\textsubscript{0.4} & \cellcolor[HTML]{ADDEC6}49.3\textsubscript{2.9} & \cellcolor[HTML]{6BC398}88.5\textsubscript{0.7} \\
SID (GBERT) & \cellcolor[HTML]{E5F5ED}15.9\textsubscript{0.5} & \cellcolor[HTML]{E5F5ED}15.5\textsubscript{1.2} & \cellcolor[HTML]{E7F6EE}14.7\textsubscript{2.3} & \cellcolor[HTML]{F3FAF7}7.5\textsubscript{0.9} & \cellcolor[HTML]{A5DBC1}53.7\textsubscript{3.0} & \cellcolor[HTML]{6CC499}87.5\textsubscript{0.6} \\
SID (mDeBERTa) & \cellcolor[HTML]{EBF7F1}12.4\textsubscript{2.0} & \cellcolor[HTML]{E3F4EB}17.1\textsubscript{1.3} & \cellcolor[HTML]{E5F5ED}15.9\textsubscript{0.9} & \cellcolor[HTML]{F7FCF9}5.3\textsubscript{1.1} & \cellcolor[HTML]{90D3B2}66.1\textsubscript{1.1} & \cellcolor[HTML]{68C296}90.0\textsubscript{0.4} \\
UD\sequential{}SID & \cellcolor[HTML]{E2F3EB}17.7\textsubscript{1.5} & \cellcolor[HTML]{DCF1E7}21.3\textsubscript{0.4} & \cellcolor[HTML]{E1F3EA}18.0\textsubscript{2.0} & \cellcolor[HTML]{F7FCFA}5.1\textsubscript{0.7} & \cellcolor[HTML]{95D4B5}63.3\textsubscript{1.7} & \cellcolor[HTML]{68C296}90.3\textsubscript{0.4} \\
UD\multitask{}SID & \cellcolor[HTML]{EEF9F4}10.2\textsubscript{0.7} & \cellcolor[HTML]{E6F5EE}14.9\textsubscript{2.9} & \cellcolor[HTML]{E8F6EF}13.8\textsubscript{1.3} & \cellcolor[HTML]{F9FDFB}3.8\textsubscript{1.0} & \cellcolor[HTML]{9ED8BC}57.8\textsubscript{2.9} & \cellcolor[HTML]{67C296}90.5\textsubscript{0.5} \\
NER\sequential{}SID & \cellcolor[HTML]{DBF1E6}21.6\textsubscript{1.1} & \cellcolor[HTML]{D6EFE2}24.9\textsubscript{3.1} & \cellcolor[HTML]{DFF2E9}19.6\textsubscript{2.0} & \cellcolor[HTML]{F2FAF6}7.9\textsubscript{1.1} & \cellcolor[HTML]{93D4B4}64.7\textsubscript{1.2} & \cellcolor[HTML]{67C295}91.0\textsubscript{0.3} \\
NER\multitask{}SID & \cellcolor[HTML]{E0F3EA}18.5\textsubscript{2.4} & \cellcolor[HTML]{D4EEE2}25.6\textsubscript{1.0} & \cellcolor[HTML]{DFF2E9}19.6\textsubscript{1.4} & \cellcolor[HTML]{EFF9F4}10.1\textsubscript{0.3} & \cellcolor[HTML]{95D4B5}63.5\textsubscript{0.9} & \cellcolor[HTML]{68C296}90.3\textsubscript{0.5} \\
MLM\sequential{}SID & \cellcolor[HTML]{F0F9F5}9.0\textsubscript{2.3} & \cellcolor[HTML]{E6F5EE}14.9\textsubscript{2.4} & \cellcolor[HTML]{EBF7F1}12.3\textsubscript{2.6} & \cellcolor[HTML]{F9FDFB}3.9\textsubscript{0.9} & \cellcolor[HTML]{9ED8BB}58.3\textsubscript{3.1} & \cellcolor[HTML]{68C296}90.1\textsubscript{0.9} \\
MLM\multitask{}SID & \cellcolor[HTML]{ECF7F2}11.9\textsubscript{1.5} & \cellcolor[HTML]{E4F4EC}16.4\textsubscript{2.7} & \cellcolor[HTML]{E7F6EE}14.6\textsubscript{1.3} & \cellcolor[HTML]{F9FDFB}3.8\textsubscript{0.0} & \cellcolor[HTML]{95D5B6}63.2\textsubscript{1.5} & \cellcolor[HTML]{65C194}91.9\textsubscript{0.3} \\
UD\sequential{}NER\sequential{}SID & \cellcolor[HTML]{DBF1E6}21.5\textsubscript{0.2} & \cellcolor[HTML]{D6EFE2}24.9\textsubscript{3.7} & \cellcolor[HTML]{DBF1E6}21.5\textsubscript{3.2} & \cellcolor[HTML]{F3FAF7}7.4\textsubscript{0.6} & \cellcolor[HTML]{92D3B3}65.1\textsubscript{3.5} & \cellcolor[HTML]{67C295}90.7\textsubscript{1.0} \\
UD\multitask{}NER\sequential{}SID & \cellcolor[HTML]{DCF1E7}21.1\textsubscript{1.6} & \cellcolor[HTML]{D5EEE2}25.5\textsubscript{1.9} & \cellcolor[HTML]{DDF2E8}20.4\textsubscript{1.6} & \cellcolor[HTML]{F3FBF7}7.3\textsubscript{0.3} & \cellcolor[HTML]{95D5B6}63.1\textsubscript{1.4} & \cellcolor[HTML]{68C296}89.9\textsubscript{0.8} \\
UD\multitask{}NER\multitask{}SID & \cellcolor[HTML]{E8F6EF}14.0\textsubscript{1.2} & \cellcolor[HTML]{E3F4EC}16.7\textsubscript{2.6} & \cellcolor[HTML]{E7F6EE}14.7\textsubscript{2.3} & \cellcolor[HTML]{F6FBF9}5.9\textsubscript{2.3} & \cellcolor[HTML]{9FD8BC}57.4\textsubscript{3.2} & \cellcolor[HTML]{67C295}90.7\textsubscript{0.5} \\
MLM\multitask{}UD\sequential{}SID & \cellcolor[HTML]{E1F3EA}18.2\textsubscript{1.7} & \cellcolor[HTML]{DCF1E7}21.2\textsubscript{1.3} & \cellcolor[HTML]{E0F3EA}18.5\textsubscript{1.8} & \cellcolor[HTML]{F7FCF9}5.3\textsubscript{0.1} & \cellcolor[HTML]{98D6B7}61.7\textsubscript{2.0} & \cellcolor[HTML]{69C397}89.5\textsubscript{1.0} \\
MLM\sequential{}NER\sequential{}SID & \cellcolor[HTML]{DFF3E9}19.1\textsubscript{1.1} & \cellcolor[HTML]{D8F0E4}23.5\textsubscript{1.1} & \cellcolor[HTML]{DDF1E7}20.7\textsubscript{0.6} & \cellcolor[HTML]{F4FBF7}7.0\textsubscript{1.3} & \cellcolor[HTML]{95D4B5}63.4\textsubscript{0.7} & \cellcolor[HTML]{68C296}90.3\textsubscript{0.5} \\
MLM\multitask{}NER\sequential{}SID & \cellcolor[HTML]{DDF2E8}20.5\textsubscript{0.2} & \cellcolor[HTML]{D4EEE1}25.7\textsubscript{2.4} & \cellcolor[HTML]{DAF0E5}22.6\textsubscript{1.6} & \cellcolor[HTML]{F0F9F5}9.2\textsubscript{0.5} & \cellcolor[HTML]{91D3B3}65.5\textsubscript{0.7} & \cellcolor[HTML]{69C397}89.5\textsubscript{1.1}\\ 
MLM\multitask{}NER\multitask{}SID & \cellcolor[HTML]{DEF2E8}20.1\textsubscript{0.6} & \cellcolor[HTML]{D4EEE2}25.6\textsubscript{2.0} & \cellcolor[HTML]{DCF1E7}21.3\textsubscript{1.2} & \cellcolor[HTML]{EEF8F3}10.3\textsubscript{1.4} & \cellcolor[HTML]{92D3B4}64.9\textsubscript{1.6} & \cellcolor[HTML]{66C195}91.2\textsubscript{0.7} \\
MLM\multitask{}UD\multitask{}NER\multitask{}SID & \cellcolor[HTML]{E6F5EE}15.1\textsubscript{1.3} & \cellcolor[HTML]{DFF3E9}19.1\textsubscript{2.3} & \cellcolor[HTML]{E3F4EC}16.7\textsubscript{1.6} & \cellcolor[HTML]{F3FAF7}7.4\textsubscript{1.1} & \cellcolor[HTML]{9DD7BB}58.9\textsubscript{1.5} & \cellcolor[HTML]{66C295}91.1\textsubscript{0.5} \\
\bottomrule
\end{tabular}
\caption{\textbf{Proportions of fully correctly classified sentences (slots and intents) in the three Bavarian dialects, Swiss German, German, and English.}
We show mean scores (in~\%) over three random seeds, with standard deviations in subscripts.
}
\label{tab:results-detailed-exactmatch}
\end{table*}

We include tables with detailed results for the Bavarian dialects, in addition to results for Swiss German, German, and English.
Table~\ref{tab:results-detailed-intents} shows the intent classification scores, Table~\ref{tab:results-detailed-slots} the slot detection scores, and Table~\ref{tab:results-detailed-exactmatch} for fully correct classifications (slots and intents).

\section{Auxiliary Task Scores}
\label{sec:appendix-auxiliary-dev}
\begin{table*}
\newlength{\stdevsep}
\setlength{\stdevsep}{4pt}
\setlength{\tabcolsep}{3pt}
\centering
\begin{tabular}{@{}lr@{\hspace{\stdevsep}}rr@{\hspace{\stdevsep}}rr@{\hspace{\stdevsep}}rr@{\hspace{\stdevsep}}rcr@{\hspace{\stdevsep}}rr@{\hspace{\stdevsep}}r@{}}
\toprule
 & \multicolumn{8}{c}{Dev scores} &  & \multicolumn{4}{c}{Test scores} \\ \cmidrule(lr){2-9} \cmidrule(l){11-14} 
 & \multicolumn{2}{@{}c@{}}{LAS $\uparrow$} & \multicolumn{2}{@{}c@{}}{POS $\uparrow$} & \multicolumn{2}{@{}c@{}}{NER $\uparrow$} & \multicolumn{2}{@{}c@{}}{PPL $\downarrow$} & & \multicolumn{2}{@{}c@{}}{Intents $\uparrow$} & \multicolumn{2}{@{}c@{}}{Slots $\uparrow$} \\ \midrule
SID (mDeBERTa) &  &  &  &  &  &  &  &  &  & \cellcolor[HTML]{FFFFFF}73.5 & \textsubscript{6.6} & \cellcolor[HTML]{FFFFFF}45.3 & \textsubscript{2.0} \\
UD\sequential{}SID & {\cellcolor[HTML]{5C98CF}{\color[HTML]{FFFFFF} 74.5}} & {\textsubscript{0.6}} & {\cellcolor[HTML]{EDF4FA}84.8} & {\textsubscript{0.3}} &  &  &  &  &  & \cellcolor[HTML]{F6FCF9}73.8 & \textsubscript{7.3} & \cellcolor[HTML]{B9E3CE}49.3 & \textsubscript{2.2} \\
UD\multitask{}SID & {\cellcolor[HTML]{FFFFFF}58.8} & {\textsubscript{9.8}} & {\cellcolor[HTML]{FFFFFF}84.3} & {\textsubscript{0.7}} &  &  &  &  &  & \cellcolor[HTML]{E67C73}48.9 & \textsubscript{7.6} & \cellcolor[HTML]{F0B5B0}42.1 & \textsubscript{4.5} \\
NER\sequential{}SID &  &  &  &  & \cellcolor[HTML]{3D85C6}{\color[HTML]{FFFFFF} 73.5} & \textsubscript{1.3} &  &  &  & \cellcolor[HTML]{99D6B8}76.6 & \textsubscript{6.8} & \cellcolor[HTML]{76C89F}53.1 & \textsubscript{2.7} \\
NER\multitask{}SID &  &  &  &  & \cellcolor[HTML]{EAF2F9}65.0 & \textsubscript{1.0} &  &  &  & \cellcolor[HTML]{A7DBC2}76.2 & \textsubscript{7.3} & \cellcolor[HTML]{69C397}53.8 & \textsubscript{2.0} \\
MLM\sequential{}SID &  &  &  &  &  &  & {\cellcolor[HTML]{FFFFFF}436.4} & {\textsubscript{22.2}} &  & \cellcolor[HTML]{FDF5F5}71.8 & \textsubscript{7.1} & \cellcolor[HTML]{E67C73}39.6 & \textsubscript{3.8} \\
MLM\multitask{}SID &  &  &  &  &  &  & {\cellcolor[HTML]{458AC8}{\color[HTML]{FFFFFF} 5.8}} & {\textsubscript{0.3}} &  & \cellcolor[HTML]{FDF6F5}71.9 & \textsubscript{7.6} & \cellcolor[HTML]{FBEEED}44.6 & \textsubscript{2.5} \\
UD\sequential{}NER\sequential{}SID & {\cellcolor[HTML]{5594CD}{\color[HTML]{FFFFFF} 75.2}} & {\textsubscript{0.7}} & {\cellcolor[HTML]{D4E4F3}85.6} & {\textsubscript{0.9}} & \cellcolor[HTML]{4F90CC}{\color[HTML]{FFFFFF} 72.6} & \textsubscript{0.4} &  &  &  & \cellcolor[HTML]{61BF91}78.3 & \textsubscript{6.4} & \cellcolor[HTML]{60BF91}54.3 & \textsubscript{3.3} \\
UD\multitask{}NER\sequential{}SID & {\cellcolor[HTML]{3D85C6}{\color[HTML]{FFFFFF} 77.4}} & {\textsubscript{9.1}} & {\cellcolor[HTML]{3D85C6}{\color[HTML]{FFFFFF} 90.0}} & {\textsubscript{0.1}} & \cellcolor[HTML]{679FD3}{\color[HTML]{FFFFFF} 71.4} & \textsubscript{1.0} &  &  &  & \cellcolor[HTML]{5EBE8F}78.4 & \textsubscript{5.8} & \cellcolor[HTML]{6BC398}53.7 & \textsubscript{1.9} \\
UD\multitask{}NER\multitask{}SID & {\cellcolor[HTML]{A8C8E6}67.2} & {\textsubscript{9.4}} & {\cellcolor[HTML]{B7D2EA}86.4} & {\textsubscript{0.2}} & \cellcolor[HTML]{FFFFFF}63.9 & \textsubscript{0.7} &  &  &  & \cellcolor[HTML]{F3C4C0}62.6 & \textsubscript{6.2} & \cellcolor[HTML]{FCF1F0}44.7 & \textsubscript{4.6} \\
MLM\multitask{}UD\sequential{}SID & {\cellcolor[HTML]{5192CC}{\color[HTML]{FFFFFF} 75.5}} & {\textsubscript{4.0}} & {\cellcolor[HTML]{C2D9ED}86.1} & {\textsubscript{0.7}} &  &  & {\cellcolor[HTML]{BCD7EF}44.8} & {\textsubscript{1.3}} &  & \cellcolor[HTML]{EFF9F4}74.0 & \textsubscript{6.0} & \cellcolor[HTML]{BBE4CF}49.2 & \textsubscript{2.7} \\
MLM\sequential{}NER\sequential{}SID &  &  &  &  & \cellcolor[HTML]{81B0DA}{\color[HTML]{FFFFFF} 70.1} & \textsubscript{2.6} & {\cellcolor[HTML]{FFFFFF}436.4} & {\textsubscript{22.2}} &  & \cellcolor[HTML]{93D3B4}76.8 & \textsubscript{7.4} & \cellcolor[HTML]{92D3B3}51.5 & \textsubscript{2.2} \\
MLM\multitask{}NER\sequential{}SID &  &  &  &  & \cellcolor[HTML]{498DCA}{\color[HTML]{FFFFFF} 72.9} & \textsubscript{0.6} & {\cellcolor[HTML]{669FD3}{\color[HTML]{FFFFFF} 7.0}} & {\textsubscript{1.8}} &  & \cellcolor[HTML]{57BB8A}78.6 & \textsubscript{7.2} & \cellcolor[HTML]{6BC398}53.7 & \textsubscript{2.3} \\
MLM\multitask{}NER\multitask{}SID &  &  &  &  & \cellcolor[HTML]{CFE1F1}66.3 & \textsubscript{0.3} & {\cellcolor[HTML]{4389C8}{\color[HTML]{FFFFFF} 5.7}} & {\textsubscript{0.4}} &  & \cellcolor[HTML]{6FC59B}77.9 & \textsubscript{7.5} & \cellcolor[HTML]{57BB8A}54.8 & \textsubscript{1.7} \\
MLM\multitask{}UD\multitask{}NER\multitask{}SID & {\cellcolor[HTML]{6DA4D5}{\color[HTML]{FFFFFF} 72.8}} & {\textsubscript{1.8}} & {\cellcolor[HTML]{B1CEE8}86.6} & {\textsubscript{0.7}} & \cellcolor[HTML]{FEFFFF}64.0 & \textsubscript{0.6} & {\cellcolor[HTML]{3D85C6}{\color[HTML]{FFFFFF} 5.5}} & {\textsubscript{0.2}} &  & \cellcolor[HTML]{EFACA6}58.0 & \textsubscript{7.9} & \cellcolor[HTML]{C3E7D6}48.7 & \textsubscript{2.4} \\
\bottomrule
\end{tabular}
\caption{\textbf{Development set scores for the auxiliary tasks} 
(LAS\,=\,labelled attachment score; 
POS\,=\,POS tagging accuracy;
NER\,=\,NER span F1;
PPL\,=\,masked token perplexity).
For context, we also show the intent accuracy and slot-filling span F1 score on the Bavarian test sets.
All scores are averaged over three runs, the SID scores are additionally averaged over the three Bavarian test sets.
Subscript numbers are standard deviations.
Darker background colours indicate better results for the auxiliary task scores. For the SID results, green cell backgrounds indicate better results than the baseline, and red worse results.
}
\label{tab:aux-dev-results}
\end{table*}

Table~\ref{tab:aux-dev-results} shows the scores on the development sets of the auxiliary tasks.

\section{Additional Bavarian Test Sets}
\label{sec:appendix-additional-data}
\begin{table*}
\centering
\adjustbox{max width=\textwidth}{
\begin{tabular}{@{}lrrrlrrrlrrr}
\toprule
 & \multicolumn{3}{c}{Intents (acc., in \%)} &  & \multicolumn{3}{c}{Slots (span F1, in \%)} &  & \multicolumn{3}{c}{Fully correct (in \%)} \\ \cmidrule(lr){2-4} \cmidrule(lr){6-8} \cmidrule(l){10-12} 
 & \multicolumn{1}{c}{de-ba} & \multicolumn{1}{c}{nat.} & \multicolumn{1}{c}{MAS.} &  & \multicolumn{1}{c}{de-ba} & \multicolumn{1}{c}{nat.} & \multicolumn{1}{c}{MAS.} &  & \multicolumn{1}{c}{de-ba} & \multicolumn{1}{c}{nat.}  & \multicolumn{1}{c}{MAS.}\\ \midrule
SID (mDeBERTa) & \cellcolor[HTML]{7DCBA5}77.7\textsubscript{0.7} & \cellcolor[HTML]{99D6B8}60.8\textsubscript{1.4} & \cellcolor[HTML]{A3DABF}55.2\textsubscript{3.5} &  & \cellcolor[HTML]{B1E0C9}46.7\textsubscript{2.0} & \cellcolor[HTML]{CAEADA}31.7\textsubscript{2.3} & \cellcolor[HTML]{DAF0E6}22.1\textsubscript{1.4} &  & \cellcolor[HTML]{E3F4EB}17.1\textsubscript{1.3} & \cellcolor[HTML]{EAF7F0}12.9\textsubscript{1.6} & \cellcolor[HTML]{F4FBF8}6.7\textsubscript{1.0} \\
MLM\multitask{}NER\multitask{}SID & \cellcolor[HTML]{71C69C}84.7\textsubscript{0.5} & \cellcolor[HTML]{99D6B8}61.0\textsubscript{3.9} & \cellcolor[HTML]{A5DBC1}53.8\textsubscript{2.5} &  & \cellcolor[HTML]{A0D9BD}56.6\textsubscript{0.3} & \cellcolor[HTML]{B8E3CE}42.3\textsubscript{2.1} & \cellcolor[HTML]{CDEBDC}30.3\textsubscript{0.9} &  & \cellcolor[HTML]{D4EEE2}25.6\textsubscript{2.0} & \cellcolor[HTML]{DDF2E8}20.3\textsubscript{1.3} & \cellcolor[HTML]{EEF8F3}10.6\textsubscript{0.8} \\
MLM\multitask{}NER\sequential{}SID & \cellcolor[HTML]{6FC59B}85.8\textsubscript{1.3} & \cellcolor[HTML]{8ED2B1}67.5\textsubscript{1.3} & \cellcolor[HTML]{9BD7B9}60.1\textsubscript{1.2} &  & \cellcolor[HTML]{A1D9BD}56.5\textsubscript{0.9} & \cellcolor[HTML]{BAE3CF}41.4\textsubscript{2.5} & \cellcolor[HTML]{CAEADA}32.0\textsubscript{1.4} &  & \cellcolor[HTML]{D4EEE1}25.7\textsubscript{2.4} & \cellcolor[HTML]{DEF2E8}20.2\textsubscript{1.0} & \cellcolor[HTML]{EBF7F1}12.4\textsubscript{0.4} \\ \bottomrule
\end{tabular}
}
\caption{\textbf{Performances on different data sets with dialects from Upper Bavaria:}
xSID (de-ba), naturalistic data (nat.), and a translated subset of MASSIVE (MAS.).
The scores are averaged across three random seeds, with standard deviations in subscripts.}
\label{tab:extra-data-massive-natural}
\end{table*}

Table~\ref{tab:extra-data-massive-natural} shows results on the \deba{} dataset in addition to other data in the same dialect (or dialects spoken in the same region).

\section{Additional Examples}
\label{sec:appendix-examples}

\begin{table*}
\adjustbox{max width=\textwidth}{%
\begin{tabular}{@{}llllllllll@{}}
\multicolumn{8}{@{}l@{}}{\dialectname{\munichdata} \qquad (intent: \gold{reminder/set\_reminder}, predicted: \predwrong{weather/find})}\\
Erinnad & mi & dass & i & morgn & papia & tiacha & im & lodn & hoi\\[\shortrow]
\gloss{Remind} & \gloss{me} & \gloss{that} & \gloss{I} & \gloss{tomorrow} & \gloss{paper} & \gloss{towels} & \gloss{in.the} & \gloss{store} & \gloss{fetch.\textsc{1sg}}\\
\gold{O} & \gold{O} & \gold{O} & \gold{O} & \gold{B--datet.} & \gold{\begin{tabular}[t]{@{}l@{}}B--rem./\\ todo\end{tabular}} & \gold{\begin{tabular}[t]{@{}l@{}}I--rem./\\ todo\end{tabular}} & \gold{\begin{tabular}[t]{@{}l@{}}I--rem./\\ todo\end{tabular}} & \gold{\begin{tabular}[t]{@{}l@{}}I--rem./\\ todo\end{tabular}} & \gold{\begin{tabular}[t]{@{}l@{}}I--rem./\\ todo\end{tabular}} \\
\predcorrect{O} & \predcorrect{O} & \predcorrect{O} & \predcorrect{O} & \predcorrect{B--datet.} & \predcorrect{\begin{tabular}[t]{@{}l@{}}B--rem./\\ todo\end{tabular}} & \predwrong{O} & \predwrong{O} & \predwrong{O} & \predwrong{O}\\
\end{tabular}}

\vspace{\baselineskip}
\adjustbox{max width=\textwidth}{%
\begin{tabular}{@{}lllllllll@{}}
\multicolumn{8}{@{}l@{}}{\dialectname{de-ba} \qquad (intent: \gold{reminder/set\_reminder}, predicted: \predcorrect{reminder/set\_reminder})}\\
Erinner & mi & moang & Papiertaschentücher & im & Ladn & zum & hoin \\[\shortrow]
\gloss{Remind} & \gloss{me} & \gloss{tomorrow} & \gloss{paper towels} & \gloss{in.the} & \gloss{store} & \gloss{\textsc{part+det}} & \gloss{fetch.\textsc{inf} (nominalized)} \\
\gold{O} & \gold{O} & \gold{B-datet.} & \gold{B--rem./todo} & \gold{\begin{tabular}[t]{@{}l@{}}I--rem./\\ todo\end{tabular}} & \gold{\begin{tabular}[t]{@{}l@{}}I--rem./\\ todo\end{tabular}} & \gold{\begin{tabular}[t]{@{}l@{}}I--rem./\\ todo\end{tabular}} & \gold{I--rem./todo}\\
\predcorrect{O} & \predcorrect{O} & \predwrong{O} & \predcorrect{B--rem./todo} & \begin{tabular}[t]{@{}l@{}}\textcolor{darkgreen}{I--rem./}\\\predcorrect{ todo}\end{tabular} & \begin{tabular}[t]{@{}l@{}}\textcolor{darkgreen}{I--rem./}\\\predcorrect{ todo}\end{tabular} & \begin{tabular}[t]{@{}l@{}}\textcolor{darkgreen}{I--rem./}\\\predcorrect{ todo}\end{tabular} & \predcorrect{I--rem./todo} \\
\end{tabular}}

\vspace{\baselineskip}
\adjustbox{max width=\textwidth}{%
\vspace{\baselineskip}
\begin{tabular}{@{}lllllllll@{}}
\multicolumn{8}{@{}l@{}}{\dialectname{de-st} \qquad (intent: \gold{reminder/set\_reminder}, predicted: \predcorrect{reminder/set\_reminder})}\\
Erinner & mi & morgn & in & Gscheft & a & Küchnrolle & zi & kafn  \\[\shortrow]
\gloss{Remind} & \gloss{me} & \gloss{tomorrow} & \gloss{in(.the)} & \gloss{store} & \gloss{a} & \gloss{kitchen roll} & \gloss{to} & \gloss{buy.\textsc{inf}} \\
\gold{O} & \gold{O} & \gold{B-datet.} & \gold{\begin{tabular}[t]{@{}l@{}}B-rem./\\ todo\end{tabular}} & \gold{\begin{tabular}[t]{@{}l@{}}I--rem./\\ todo\end{tabular}} & \gold{\begin{tabular}[t]{@{}l@{}}I--rem./\\ todo\end{tabular}} & \gold{\begin{tabular}[t]{@{}l@{}}I--rem./\\ todo\end{tabular}} & \gold{\begin{tabular}[t]{@{}l@{}}I--rem./\\ todo\end{tabular}} & \gold{I--rem./todo}\\
\predcorrect{O} & \predcorrect{O} & \predcorrect{B-datet.} & \predwrong{O} & \begin{tabular}[t]{@{}l@{}}\textcolor{darkgreen}{I--rem./}\\\predcorrect{ todo}\end{tabular} & \begin{tabular}[t]{@{}l@{}}\textcolor{darkgreen}{I--rem./}\\\predcorrect{ todo}\end{tabular} & \begin{tabular}[t]{@{}l@{}}\textcolor{darkgreen}{I--rem./}\\\predcorrect{ todo}\end{tabular} & \begin{tabular}[t]{@{}l@{}}\textcolor{darkgreen}{I--rem./}\\\predcorrect{ todo}\end{tabular} & \predcorrect{I--rem./todo} \\
\end{tabular}}
\caption{\textbf{Translations of ``Remind me to get paper towels at the store tomorrow'' into Bavarian dialects with \gold{gold-standard} and (\predcorrect{correctly} or \predwrong{incorrectly}) predicted annotations.}
Note the different syntactic structures for expressing the infinitive or subordinated phrase, the different translations used for ``store'' and ``paper towels'' (and the different order in which they are mentioned), and the spelling differences (e.g., for ``tomorrow'').
The predictions are by the overall best-performing model, MLM\multitask{}NER\sequential{}SID, with the same random seed.
Abbreviated slots: datet.~=~datetime, rem.~=~reminder.
}
\label{tab:reminder-get-paper}
\end{table*}

Table~\ref{tab:reminder-get-paper} provides another example for translation (and prediction) differences between the Bavarian dialects.

\end{document}